\documentclass[acmsmall,screen]{acmart}

\setcopyright{none}
\copyrightyear{2023}
\acmYear{2023}
\acmDOI{}

\acmVolume{0}
\acmNumber{0}
\acmArticle{0}

\usepackage{booktabs}
\usepackage{xcolor}
\usepackage{mdframed}
\usepackage{environ}
\usepackage{comment}
\theoremstyle{definition}
\newtheorem{definition}{Definition}

\usepackage{adjustbox}
\usepackage{multirow}

\theoremstyle{remark}

\newtheorem{example}{Example}

\newcommand{\mfact}[3]{\ensuremath{\langle{}#1, #2, #3\rangle}}
\newcommand{\fact}[3]{\ensuremath{\langle{}}\textit{#1, #2, #3}\ensuremath{\rangle}}

\NewEnviron{revision}{\textcolor{black}{\BODY}}  

\newmdenv[
  topline=false,
  bottomline=false,
  rightline=false,
  linecolor=lightgray,
  linewidth=4pt,
  innertopmargin=1pt,
  innerbottommargin=1pt,
  innerleftmargin=5pt,
  innerrightmargin=0pt,
  skipabove=5pt,
  skipbelow=5pt
]{siderules}

\renewenvironment{example}{\begin{siderules} \textbf{Example:}}{\end{siderules}}

\renewcommand{\paragraph}[1]{\smallskip\noindent\textbf{#1\mbox{\ }}}

\newcommand{\problem}[3]{
\begin{definition}#1\\
\textbf{Input:} #2\\
\textbf{Task:} #3
\end{definition}
}

\begin{document}

\title[Completeness, Recall and Negation]{Completeness, Recall, and Negation\\ in Open-World Knowledge Bases: A Survey}


\author{Simon Razniewski}
\email{simon.razniewski@de.bosch.com}
\affiliation{\institution{Bosch Center for AI}\country{Germany}}
\author{Hiba Arnaout}
\email{harnaout@mpi-inf.mpg.de}
\affiliation{\institution{Max Planck Institute for Informatics}\country{Germany}}
\author{Shrestha Ghosh}
\email{ghoshs@mpi-inf.mpg.de}
\affiliation{\institution{Max Planck Institute for Informatics}\country{Germany}}
\author{Fabian Suchanek}
\affiliation{%
  \institution{Telecom Paris, Institut Polytechnique de Paris}
  \country{France}
}
\email{suchanek@telecom-paris.fr}

\begin{abstract}
General-purpose knowledge bases (KBs) are a cornerstone of knowledge-centric AI. Many of them are constructed pragmatically from web sources, and are thus far from complete. This poses challenges for the consumption as well as the curation of their content.
While several surveys target the problem of completing incomplete KBs, the first problem is arguably to know whether and where the KB is incomplete in the first place, and to which degree.

In this survey, we discuss how knowledge about completeness, recall, and negation in KBs can be expressed, extracted, and inferred. We cover (i) the logical foundations of knowledge representation and querying under partial closed-world semantics; (ii) the estimation of this information via statistical patterns; (iii) the extraction of information about recall from KBs and text; (iv) the identification of interesting negative statements; and (v) relaxed notions of relative recall. 

This survey is targeted at two types of audiences: (1) practitioners who are interested in tracking KB quality, focusing extraction efforts, and building quality-aware downstream applications; and (2) data management, knowledge base and semantic web researchers who wish to understand the state of the art of knowledge bases beyond the open-world assumption. Consequently, our survey presents both fundamental methodologies and their working, and gives practice-oriented recommendations on how to choose between different approaches for a problem at hand.
\end{abstract}




\maketitle

\section{Introduction}

\paragraph{Motivation.}
Web-scale knowledge bases (KBs) like Wikidata~\cite{wikidata}, DBpedia~\cite{dbpedia}, NELL~\cite{nell}, or YAGO~\cite{YAGO} are a cornerstone of the Semantic Web. Pragmatically constructed from web resources, they focus on representing \textit{positive} knowledge, i.e., statements that are true. However, they typically contain only a small subset of all true statements, without qualifying what that subset is. For example, a KB may contain winners of the \textit{Nobel Prize in Physics}, but it does not necessarily contain \emph{all} winners. It will not even specify whether it contains all winners or not. For example, a KB may lack the information that a specific renowned physicist won the \textit{Nobel Prize}, but this does not mean that this person did not win the award -- the data may just be lacking. Vice versa, if it is known that a specific physicist did definitively not win the \textit{Nobel Prize}, there is no way to express this in most current KBs.
A KB may also not contain a specific scientist at all -- without any indication that she or he is missing.\footnote{We refrain from giving an example of an incomplete entity in a real-world KB, because whenever we did that in previous publications, helpful readers specifically completed these entities in the KB, thereby rendering our example outdated.}

Such uncertainty about the extent of knowledge poses major challenges for the curation and application of KBs:
\begin{enumerate}
    \item \textbf{Human KB curators} need to know where the KB is incomplete, so that they can prioritize their completion efforts. For example, when working on a KB like Wikidata, with over 100M items, systematic knowledge on gaps and quality is essential in order to know where to focus limited human resources.
    \item \textbf{Automated KB construction pipelines} need this knowledge, too, in order to know how to adjust acceptance thresholds. For example, if we already have 218 Physics Nobel prize winners in the KB, and if we know that this is the expected count, further candidates should all be rejected. In contrast, if no publications of a scientist are recorded in the KB, we should accept automated extractions. 
    This is significant in particular for KB projects such as NELL, which aim to auto-complete themselves.
    \item \textbf{QA Applications} are built on top of KBs. They need to know where the data is incomplete, so as to alert end users of quality issues. For example, a query for ``\textit{the astronomer who discovered most planets}'' may return the wrong answer if \textit{Geoffrey Marcy} happens to be absent from the KB. Similarly, a KB that is used for question-answering should have awareness of when a question surpasses its knowledge~\cite{rajpurkar2018know}. This holds in particular for Boolean questions such as ``\textit{Does Harvard have a physics department?}'' (where a ``no'' could come simply from missing information), and analytical aggregate questions such as ``\textit{How many US universities have physics departments?}'' (where receiving some answer gives little clue about its correctness).
    \item \textbf{Structured entity summaries} need awareness not only of positive properties, but also of salient negatives. For example, one of the most notable properties of \textit{Switzerland} is that it is not a \textit{EU} member (despite heavy economic ties and being surrounded by EU countries). For \textit{Stephen Hawking}, a salient summary of his accolades should include that he did \textit{not} win the \#1 award in his field, the \textit{Nobel Prize in Physics}. 
    \item \textbf{Machine learning} on knowledge bases and text, in particular, for the tasks of KB link prediction and textual information extraction, needs negative training examples to build accurate models. Obtaining quality negative examples is a major hurdle when working on these tasks, and much research has focused on heuristics, for example, based on obfuscation of positive statements. Explicit negations, or negations derived from completeness metadata, could significantly impact these tasks.
\end{enumerate}    

\noindent This survey presents the methods that the recent literature has developed to address these problems. For example, several formalisms have been developed to specify the extent of the knowledge in a KB, including the possibility to make negative statements, or to specify metadata on completeness. Knowledge engineers can record that they entered \textit{all} winners of the \textit{Nobel Prize in Physics} into Wikidata, or mark prominent Physicists who did \textit{not} win it. If such a manual marking is not possible, there are also methods to infer the completeness of a KB automatically. For example, it is possible to spot phrases such as \textit{``the Nobel Prize in Physics was awarded to 218 individuals''}, and to compare this count with the number of entities in the KB. It is also possible to draw conclusions about the completeness of a KB from the growth or near-stagnation of a set of entities in the KB, from the overlap between sets of \textit{Nobel Prize} winners from different random samples, or from different KBs. One can also observe that most winners hold academic degrees, and so flag entities without an alma mater as likely incomplete. The knowledge engineer thus has a growing repository of methods at hand to tackle the problem of incompleteness, and this is what the present survey covers.

\paragraph{Focus of this survey.}
Several comprehensive surveys discuss KBs at large, in particular their construction, storage, querying, and applications \cite{hogan2021knowledge,machine-knowledge-survey,ji2021survey}. Focused works on KB quality discuss how recall can be increased, and how quality can be measured by assessing accuracy and provenance~\cite{paulheim2017knowledge,zaveri2016quality,ben2018rdf,farber2018linked}. 
Recent years have also seen new 
formalisms for describing recall and negative knowledge~\cite{darari2013completeness,ARNAOUT2021100661,mirza2018enriching}, as well as the rise of statistical and text-based methods for estimating recall~\cite{galarraga2017predicting,hopkinson2018demand,lajus2018all,luggen2019non,razniewski2019coverage,soulet2018representativeness,karagiannis2019mining,balaraman2018recoin} and deriving negative statements~\cite{karagiannis2019mining,arnaout020enriching,safavi-etal-2021-negater,arnaoutcikm2022}, with some of them collected in a systematic literature review~\cite{issa2021}. The goal of this survey is to systematize, present and analyze this topic.
The survey is intended both for theoreticians and practitioners. It intends to inform the readers about the latest advances in completeness assessment and negation,
and equip them with a repertoire of methodologies to better represent and assess the recall of specific datasets. The survey builds on content that has been presented at tutorials at VLDB'21, ISWC'21, KR'21 and WWW'22. The tutorial slides are available at \url{https://www.mpi-inf.mpg.de/knowledge-base-recall/tutorials}. 

\paragraph{Outline.} This survey is structured as follows: In Section~\ref{sec:foundations}, we start with the \textbf{foundations} of the logical framework in which KBs operate, the open world assumption (OWA),
incompleteness, the implications for query answering~\cite{reiter1981closed}, as well as the formal semantics of completeness and cardinality assertions~\cite{darari2013completeness}. 
In Section~\ref{sec:predictive}, we discuss \textbf{how recall can be estimated automatically}. We present supervised machine learning methods~\cite{galarraga2017predicting}, unsupervised techniques such as species sampling~\cite{trushkowsky2013crowdsourced,luggen2019non}, density-based estimators~\cite{lajus2018all}, statistical invariants about number distributions~\cite{soulet2018representativeness}, and linguistic theories about completeness in human conversations~\cite{razniewski2019coverage}.
In Section~\ref{sec:cardinalities}, we focus on \textbf{determining the cardinality of a predicate} (i.e., its number of objects for a given subject)  from KBs and from text. We show how cardinality information from a KBs can be identified and used to 
    assess recall~\cite{ghosh2020uncovering}, how this information can be extracted from natural language documents~\cite{mirza2018enriching}, and how this differs from mining functional cardinalities that remain invariant given any subject (for instance, all humans have one birthplace)~\cite{giacometti2019mining}.
In Section~\ref{sec:negation}, we focus on \textbf{identifying salient negations}. We show why explicit negations about KB entities are needed in open-world settings. We present methods to identify negations using inferences from the KB itself~\cite{arnaout020enriching,arnaout2021wikinegata,arnaoutcikm2022,safavi-etal-2021-negater,safavi2020generating} and methods to extract negations from various textual sources~\cite{karagiannis2019mining,arnaout020enriching}, in particular query logs. We also outline open issues such as the precision/salience trade-off and ontology modelling and maintenance.
In Section~\ref{sec:relativerecall}, we discuss \textbf{relative recall}, i.e., more relaxed notions of recall that stop short of aiming to capture all knowledge from the real world. We show how recall can be measured by extrinsic use cases like question answering and entity summarization~\cite{hopkinson2018demand,razniewski2020structured}, by comparison with open information extraction or external reference resources~\cite{gashteovski2020aligning,mishra2017domain}, and by comparison with other comparable entities inside the KB~\cite{balaraman2018recoin,ores}. 
In Section 7, we conclude with \textbf{recent topics}, \textbf{recommendations} towards making KBs recall-aware, and \textbf{open research challenges}. \begin{revision}
In particular, large language models (LLMs) have recently shaken up knowledge-centric computing.
Although completeness and recall research has yet to capitalize on these advances, we highlight several ways by which LLMs are likely to impact this area.    
\end{revision}

\section{Foundations}
\label{sec:foundations}

\subsection{Knowledge Bases}

We first introduce foundational concepts, in particular, knowledge bases, their semantics, and the notions of completeness, recall and cardinality metadata. Knowledge bases are built on three pairwise disjoint infinite sets $E$ (entities), $P$ (predicates),
and $L$ (literals). Entities refer to uniquely identifiable objects such as \textit{Marie Curie}, the \textit{Nobel Prize in Physics}, or the city of \emph{Paris}. Literals are strings or numerical values, such as \textit{dates}, \emph{weights}, and \emph{names}. Predicates (also known as relations, properties, or attributes) link an entity to another entity or to a literal. Examples are \textit{birthPlace} or \textit{has\-AtomicNumber}. 
A tuple $\mfact{s}{p}{o} \in E \times P \times (E \cup L)$ 
is called a \textbf{positive statement} (also known as triple, assertion, or fact), where $s$ is the subject, $p$ the predicate and $o$ the object~\cite{knowledge-representation}. A statement says that the subject stands in the given predicate with the object, as in \fact{Marie Curie}{won\-Award}{Nobel Prize in Physics}. 
For our purposes, we will consider also \textbf{negative statements}, which say that the subject does not stand in the given relationship with the object, as in $\neg$\fact{Stephen Hawking}{wonAward}{Nobel Prize in Physics}.
Entities can be organized into classes such as \emph{Physicists}, \emph{Cities}, or \emph{Awards}. An entity can be an \emph{instance} of a class, and this can be expressed by a triple with the predicate \emph{type}, as in \fact{Marie Curie}{type}{Physicist}. Classes can  be arranged in a subclass hierarchy. For example, physicists are scientists, and scientists are people. This, too, can be expressed by triples, using the special predicate \emph{subClassOf}: \fact{Physicist}{subClassOf}{Scientist}. Some KBs allow specifying \emph{constraints}, such as domain and range constraints on predicates, functionality constraints, or disjointness constraints between classes, usually in a formalism called Description Logic
\cite{baader2003description}. 
In our case, we are concerned mainly with the statements and with class membership, and not so much with constraints (i.e., in the terminology of Description Logic, we focus on the assertional part of knowledge of the A-box, and less on the terminological knowledge of the T-box).

\begin{definition}[Knowledge Base]
A knowledge base (or knowledge graph) $K$ is a finite set of positive statements \cite{klyne2004resource}.
\end{definition}

\begin{example}
As a running example, consider a KB with biographical data about scientists. An excerpt of this KB is shown in Table~\ref{tbl:runningex}, in a Wikidata-style layout, where the subject of all statements is shown at the top (\textit{Marie Curie}), and predicates and objects are shown in tabular form below. This KB contains a diverse set of statements, linking entities with other entities, as well as with literals, and containing multiple objects for \textit{some} predicates but not for \textit{all}.

\end{example}

\begin{table}[]
\centering
\begin{tabular}{@{}ll@{}}
\toprule
{Marie Curie} &  \\ \midrule
{birthPlace} & {Warsaw} \\
{birthYear} & {1867}\\
citizenOf & Poland \\
{advised} & {Marguerite Perey} \\
 & {Óscar Moreno} \\
 & ... \textit{(6 more names)} ...\\
{discovered} & {Polonium} \\
{wonAward} & {Nobel Prize in Physics} \\
 & {Franklin Medal} \\ \bottomrule
\end{tabular}
\caption{Running example KB.} 
\label{tbl:runningex}
\end{table}


\begin{revision}
\noindent The term ``Knowledge Graph'' (KG) is also often used as an alternative to ``Knowledge Base'' -- ever since Google popularized the term in 2012~\cite{google-kg-blog}. Although 
KGs are sometimes defined as special cases of KBs, by and large, the two terms have been used synonymously. For the purpose of this survey, we follow suit, and use them interchangeably. KBs (or KGs) haven received a considerable uptake in recent years \cite{hogan2021knowledge,machine-knowledge-survey}. From humble beginnings in dedicated communities like YAGO~\cite{YAGO}, DBpedia \cite{dbpedia}, Freebase \cite{freebase}, or Wikidata \cite{wikidata}, KBs are now standard technology at most major tech companies \cite{noy2019industry}, and widely in use beyond. Nonetheless, understanding their quality remains an enduring challenge. With this survey, we contribute a detailed overview concerning the aspects of completeness, recall and negation.

In this survey, we focus on knowledge that can be found in text, and that can reasonably be expressed in structured format in knowledge bases. Intangible knowledge, e.g., around concepts such as motion, scent, rhythm, or touch, is out of scope for this paper.
\end{revision}

\subsection{Incompleteness}

Open-world knowledge bases are inherently incomplete, i.e., they store only a subset of all true statements from the domain of interest \cite{what-do-we-actually-know}. 
Our example KB in Table~\ref{tbl:runningex}, for instance,
misses 2 other citizenships that Marie Curie held (\textit{Russia}, \textit{France}), and it does not even contain her receiving the \textit{Nobel Prize in Chemistry}.\footnote{Making her one of the two people who ever received two Nobel Prizes.} Moreover, it lacks the statement that, contrary to a common misconception, she did not find the first discovered radio-active element, \textit{Uranium}.

Such incompleteness happens for several reasons:
\begin{enumerate}
    \item \textbf{Source incompleteness:} 
    KBs are usually built from another source, either by automated methods \cite{machine-knowledge-survey}, or by human curators, who read websites, textbooks, news articles, etc., or other authoritative websites.
    These sources may themselves be incomplete. 
    \item \textbf{Access limitations:} In some cases, only a subset of the relevant documents are readily accessible to automated methods or to human curators. For example, major troves of knowledge are locked away in the Deep Web (web pages without inlinks) \cite{deepweb}, and there are printed documents that are not available in digital form.
    \item \textbf{Extraction imperfection:} Even statements that are available in digital form and accessible can be missed, because extractors are imperfect. This holds especially for automated text extraction methods~\cite{machine-knowledge-survey}: the best models achieve only 75\% in recall on the popular TACRED benchmark \cite{wang2021k}.
    \item \textbf{Resource limitations:} Human efforts are naturally bounded by available work time, and this applies to a lesser degree also to automated extractors. This is especially relevant in the long tail of knowledge, where social media content emerges continuously, at fast pace.
    \item \textbf{Intractability of materializing all true negations:} For negative assertions, another difficulty joins in: The set of true negative statements is quasi-infinite\footnote{Infinite if one considers an infinite domain, finite but intractable if one considers a finite domain, e.g., the active domain of a KB.}, and thus it is infeasible to materialize negative statements beyond a few salient ones per subject.
\end{enumerate}
We formalize the (in-)completeness of a KB by help of
a hypothetical ideal KB $K^i$. This ideal KB contains
all true statements about the domain of interest in the real world. \begin{revision} 
Such an ideal KB is not an easy concept. 
In the most naive conceptualization, $K^i$ simply contains all statements that are true in reality, and that are expressible with a given set of predicates. This makes sense for relations that are sufficiently well-defined such as “sibling” or “place of birth”. However, for other relations (such as ``hobby'') this conceptualization may be ill-defined~\cite{what-do-we-actually-know}: while one of Albert Einstein’s hobbies was playing the violin, he might have had an unclear number of other “hobbies” (such as going for a walk, or eating chocolate). Thus, it is not clear what an ideal KB should contain for this relation. The same goes for entities: is anybody with a doctoral degree a scientist? Or do we could only people employed as scientists? What if a freelancing scientist makes an important discovery? This shows that we can posit the notion of the ideal KB only for well-defined sets of relations and entities: hobbies that are pursued in a registered association, scientists who are employed at a research institution, etc. While this comes at the expense of what can be targeted, many relations and sets of entities are sufficiently well-defined to establish completeness: all countries that are members of the United Nations; all mountains higher than 1000m in a given country; all universities of a given country that can deliver a doctoral degree; etc. In what follows, we shall focus on such domains where the ideal KB $K^i$ can be reasonably established.

\end{revision}

We say that a statement $st$ from the domain of interest is \emph{true in the real world}, if $st\in K^i$, and a negative statement $\neg st$ from the domain of interest is true in the real world (``truly false''), if $st\not\in K^i$.
We say that another KB $K$ is
\emph{correct}, if $K \subseteq K^i$. Correctness is usually a core focus in KB construction \cite{paulheim2017knowledge} (also referred to as precision, truthfulness or accuracy). 
Nevertheless, KBs are of course not all correct at scale. However, since we focus on incompleteness in this survey, we will make the simplifying assumption that the KB at hand is correct.

A selection condition $\sigma$ is a statement $\mfact{s}{p}{o}$, where each component can either be instantiated, or a wildcard (``$*$''). For example, $\mfact{*}{\textit{birthPlace}}{\textit{Warsaw}}$ is a selection condition. The result of a selection condition $\sigma$ on a KB $K$ (written $\sigma(K)$) is the set of all statements in $K$ that match $\sigma$ at its non-wildcard positions. For example, $\mfact{*}{\textit{birthPlace}}{\textit{Warsaw}}$ selects all people born in Warsaw, and $\mfact{\textit{Marie Curie}}{\textit{wonAward}}{*}$ selects all awards won by Marie Curie. 


We can now proceed to define major concepts for our survey.

\begin{definition}[KB Completeness]
Given a KB $K$ and a selection condition $\sigma$, we say that $K$ is \textit{complete} for $\sigma$ if 
\[
    \sigma(K) = \sigma(K^i).
\]
\end{definition}

\begin{definition}[KB Recall]
\label{def:recall}
Given a KB $K$ and a selection condition $\sigma$, the \textit{recall} of $K$ for $\sigma$ is defined as:
\[
    \textit{Recall} = \frac{|\sigma(K)|}{|\sigma(K^i)|}.
\]
\end{definition}

\noindent It follows that KB recall is a real-valued concept, while KB completeness is a binary concept, satisfied exactly when recall equals 1.

%

\paragraph{Terminology.}
\begin{revision}
The existing literature does not 
use terminology consistently. It often uses the terms completeness and recall interchangeably, and utilizes others, such as coverage. For this survey, we strictly use \textit{completeness} for the Boolean concept of whether the result of a selection condition on the available KB equals the result of the same selection on the ideal one, and \textit{recall} for its generalization to the quantitative ratio of the two. 
Formally, a separate notion of completeness is thus superfluous, yet we find the Boolean case frequent enough to indicate it separately. 
\end{revision}

\subsection{World Semantics}
%
In data management, there are traditionally two major paradigms on how to interpret positive KBs: The \textit{closed-world Assumption} (CWA) states that statements that are not in the KB are false in the real world, i.e., $\fact{s}{p}{o}\not\in K \Rightarrow \fact{s}{p}{o}\not\in K^i$. This contrasts with the \textit{Open-world Assumption} (OWA), which states that the truth of these statements is \textit{unknown}, i.e., a statement that is not in the KB might or might not be in $K^i$. The CWA is reasonable in many limited domains, e.g., a corporate database where all employees, orders, etc.\ are known, education management, where becoming a student requires completing a formal sign-up process, or professional sports where membership in major leagues is well-established. In many other settings, however, where it is impossible, unrealistic, or not desired to have all statements and entities of a given domain, the
OWA is more appropriate. For example, it is unrealistic, and even undesired, for a KB such as YAGO to contain all people of American nationality. Therefore, most web-scale KBs operate under the OWA.

\begin{example}
Consider again \textit{Marie Curie}. Consider the query \textit{``Was Marie Curie a Polish citizen?''}. Under both the OWA and the CWA, the answer would be ``Yes'', because the statement \fact{Marie Curie}{citizenship}{Polish} is in the KB. Now consider the query ``\textit{Was Mary Curie an Australian citizen?}'' Under the CWA, given that no such statement is in the KB, the answer would be ``no''. Yet under the OWA, the statement could still be true in reality, so the answer would be ``Unknown''. Finally, consider the query for ``\textit{Nobel Peace Prize winners who are not US citizens}''. Under the CWA, that query would return Marie Curie and many others. Under the OWA, this query would have to return the \textit{empty set}, because for any winner, the KB could just be missing US citizenship.
\end{example}

\noindent
The OWA and the CWA represent extreme stances, and both have severe shortcomings. The OWA makes it impossible to answer ``No'' to any query, while the CWA produces the answer ``No'' also in cases where the KB should refrain from taking a stance. Intermediate settings are thus needed, referred to as \textit{Partial-closed World Assumptions} (PCWA) \cite{motro,denecker2008towards}, where some parts of the KB are treated under closed-world semantics, others under open-world semantics. One instantiation of the PCWA is the \textit{partial completeness assumption} (PCA) \cite{amie,dong2014data}. The PCA asserts that if a subject has \textit{at least one} object for a given predicate, then there are no other objects in the real world beyond those that are in the KB.  
For example, if the KB knows only one award for \textit{Marie Curie}, then we assume that she won 
no others.
However, if no pet is given, we assume nothing about her number of pets: she still might have had pets or not. Empirically, the PCA has been found to be frequently satisfied in KBs (see Section~\ref{sec:recall-facts} - paragraph ``Weak Signals''). Still, the PCA is a generic assumption, and does not allow the specification of individual regions of completeness/incompleteness.

\subsection{Completeness and Cardinality Metadata} 
We next introduce two kinds of metadata assertions that can be used to specify areas of the KB that are complete
~\cite{motro,razniewski2011completeness}. 

\begin{definition}[Completeness Assertion]
Given an available KB $K$, and an ideal KB $K^i$, a completeness assertion \cite{motro} is a selection condition $\sigma$ for tuples that are completely recorded.
Formally, such an assertion is satisfied, if $\sigma(K^i)=\sigma(K)$.
\end{definition}
\newcommand{\complete}[1]{\textit{IsComplete}($#1$)}
\begin{example}
The completeness assertion $\mfact{\textit{Marie Curie}}{\textit{advised}}{*}$ specifies that all advisee relations for \textit{Marie Curie} are in the KB. Given the running example from Table~\ref{tbl:runningex}, this would mean that Curie advised no one else than Marguerite Perey, Óscar Moreno, and the other 6 listed individuals. 
\end{example}

\begin{figure}[t]
    \centering
    \includegraphics[width=\columnwidth]{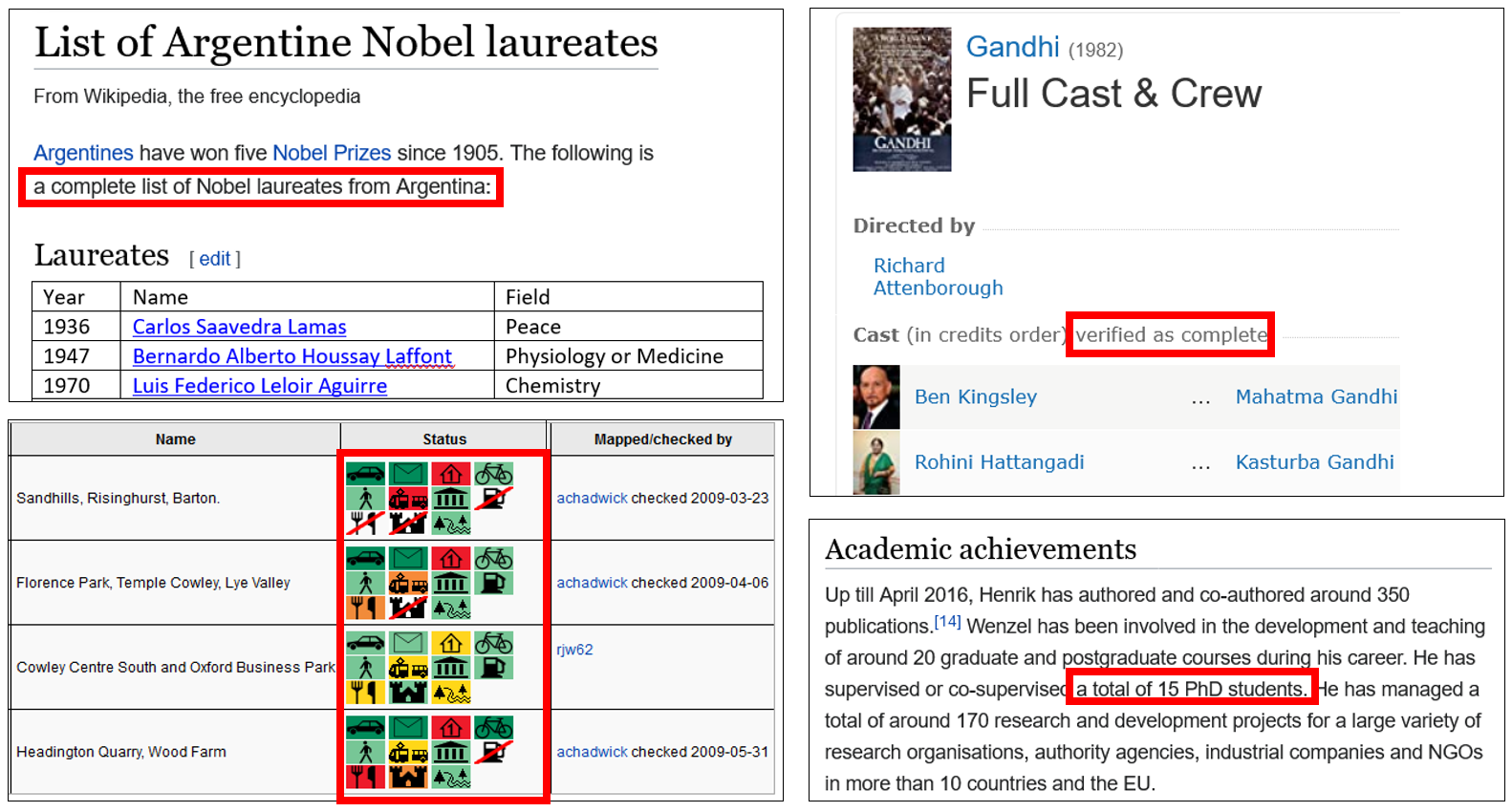}
    \caption[Caption for statement examples]{Examples of completeness and cardinality assertions on the web. Top left and bottom right: Wikipedia. Top right: IMDb. Bottom left: Openstreetmap.\footnotemark}
    \label{fig:statements-samples}
\end{figure}

\footnotetext{Image sources: \url{https://wiki.openstreetmap.org/w/index.php?title=Abingdon&oldid=471369}, \url{https://www.imdb.com/title/tt0083987/fullcredits?ref_=tt_ov_st_sm}, \url{https://en.wikipedia.org/wiki/List_of_Argentine_Nobel_laureates} and \url{https://en.wikipedia.org/wiki/Henrik_Wenzel}} 

\noindent
Completeness assertions can be naturally extended to select-project-join queries \cite{motro}. For example, one would use a join to specify that the KB knows all the advisees of Physics Nobel Prize winners from the UK. Projections are also possible, e.g., to assert completeness of all subjects that have a VIAF ID, 
but without recording the ID itself.
Both joins and projections come with subtleties, particularly whether subqueries are evaluated on the available KB, or on the conceptual ideal resource $K^i$. Similarly, projections allow to collapse multiplicities of join queries, and completeness assertions on resulting sets or bags carry different semantics (\textit{Does the KB allow to reconstruct all Nobel Prize winners, or does it also allow to reconstruct \textit{how often} each won the award?)}. We refer the reader to \cite{razniewski2011completeness} for further detail.

Various human-curated resources allow specifying completeness assertions in the spirit defined above, as shown in Figure~\ref{fig:statements-samples} (top). On the top left, one can see an assertion about the completeness of Argentinian Nobel laureates. This is an example of a typical join filter: the predicate \textit{wonAward} needs to be joined with the predicate \textit{citizenOf}. On the top right, we see a simpler assertion about cast members of a Movie. This corresponds to a selection like $\mfact{\textit{Gandhi}}{\textit{starsIn}}{*}$. 

Knowledge bases often give classes special importance (identified via the ``type'' predicate), and although both can be modelled with the same completeness assertions, one can thus divide completeness further into entity completeness (as in ``Do we have all physicists?'') and statement completeness (as in ``Do we have all awards for each physicist?''). This distinction has advantages, because the methods to tackle these types of incompleteness are different, as we shall see in Section~\ref{sec:predictive}. However, the separation is often not clear-cut, as entities are frequently identified by statements. For example, \textit{French cities} are identified by the selection \fact{*}{locatedIn}{France}. Therefore, our definition above makes no formal distinction, although works discussed in the following sections may pragmatically focus on one of the two categories of completeness.

\paragraph{Cardinality assertions}$\!\!$ 
are inspired by number restrictions from Description Logics \cite{HollunderBaader-KR-91}. They take the following form:

\begin{definition}[Cardinality Assertion]
A cardinality assertion is an assertion of the form $|\sigma(K^i)|=n$, where $\sigma$ is a selection condition, and $n$ is a natural number. It specifies the number of tuples in $K^i$ that satisfy a certain property. 
\end{definition}


\begin{example}
\begin{revision}
Continuing with the selection condition $\sigma=\mfact{\textit{Marie Curie}}{\textit{wonAward}}{*}$, the cardinality assertion $|\sigma(K^i)|=37$ expresses that in reality, \textit{Marie Curie} has received a total of 37 awards. 
\end{revision}
\end{example}    


\noindent Cardinality assertions can be used to infer recall for a given selection condition by computing the division from Definition~\ref{def:recall}, $\frac{|\sigma(K)|}{n}$. Moreover, in the special case where $|\sigma(K)|=n$, they can be used to infer completeness.

\begin{example}
Assume that we know that in reality \textit{Marie Curie} has received a total of 37 awards, i.e., $|\sigma(K^i)|=37$. 
In the running example from Table~\ref{tbl:runningex}, for the selection condition of Marie Curie's awards, $|\sigma(K)|=2$. Based on Definition~\ref{def:recall}, this would mean that on her awards, the KB has a recall of $\frac{2}{37}\approx 5.4\%$. 
\end{example}

\noindent Cardinality assertions are also found in web resources, as shown in Figure~\ref{fig:statements-samples} (bottom right), where a Wikipedia article mentions the number of PhD students advised by a certain researcher. The snippet at the bottom left contains no explicit numerals, but carries a similar spirit of recording to which degree information on districts of a city is recorded in Openstreetmap.

Completeness and cardinality assertions are closely related: cardinality assertions allow to evaluate completeness, while completeness assertions establish cardinalities. If we know that Marie Curie advised 8 PhD students, we can infer that the KB in Fig.~\ref{fig:statements-samples} is complete. If we know that she held 3 different citizenships, we can infer the KB is incomplete. If we know that the KB is complete, we can infer that the true cardinality of PhD students she advised is 7. Completeness, in turn, enables establishing negation. If we know that the list of Marie Curie's advisees is complete, we can infer that she did not advise Jean Becquerel, Louis de Broglie, etc. 

\paragraph{Reasoning with completeness metadata.} 
Given complex KB queries and metadata about completeness, a natural question is to deduce whether and which parts of the query answer are implied to be complete (or in the case of negation operations, correct). This problem has received extensive attention in database research \cite{motro,levy1996obtaining,denecker2008towards,razniewski2011completeness,lang2014partial,savkovic2013complete,razniewski2015identifying}, as well as knowledge base research \cite{darari2013completeness,darari2018completeness}. Approaches either take a deductive route (a query result is complete, if certain conditions are met) \cite{levy1996obtaining,razniewski2011completeness,darari2013completeness}, or inductively propagate metadata through query operators \cite{motro,lang2014partial,razniewski2015identifying}, with computational complexity mirroring or exceeding the complexity of query answering.

Completeness assertions may also be inferrable via constraints. Most notably, if a predicate is asserted to be functional, then, per subject, presence of one object indicates completeness. For example, if one birth place for Marie Curie is present, then birth places of Marie Curie are complete. Cardinality constraints, like \textit{every person has two parents}, naturally extend this idea, and are in turn further extended by the cardinality assertions, which make this subject-specific.

\section{Predictive Recall Assessment}
\label{sec:predictive}

In this section, we deal with approaches that can automatically determine the \emph{recall} of a KB, i.e., the proportion of the domain of interest of the real world that the KB covers. 
We consider the recall of entities (Section~\ref{sec:recall-entities}) and the recall of statements (Section~\ref{sec:recall-facts}).

\subsection{Recall of Entities}\label{sec:recall-entities}

To estimate to what degree a KB is ``complete'', we would first need to know how many entities 
from the domain of interest are missing. We can formalize this problem as follows:

\problem{Missing Entities Problem}{A class $C$ with some instances}{Determine how many instances of $C$ are missing compared to $K^i$
.}

\noindent
This definition is much less clear-cut than one would hope~\cite{what-do-we-actually-know,vagueness}: is anyone with a doctoral degree a \emph{scientist}? what is the total number of \textit{cathedrals} in a country if some have been destroyed or rebuilt? what is the total number of \textit{islands} of a country (do we also count \textit{islets} and \textit{rocks})? what is the total number of inhabitants of a country (do we also count deceased people, do we count only famous people)? Hence, in what follows, we will restrict ourselves to crisp and well-defined classes such as the countries recognized by the UN as of 2022, mountains in a certain country that are taller than 1000m, etc. Let us now look at various methods to address the Missing Entities Problem.
\begin{figure}[t]
    \includegraphics[width=0.9\textwidth]{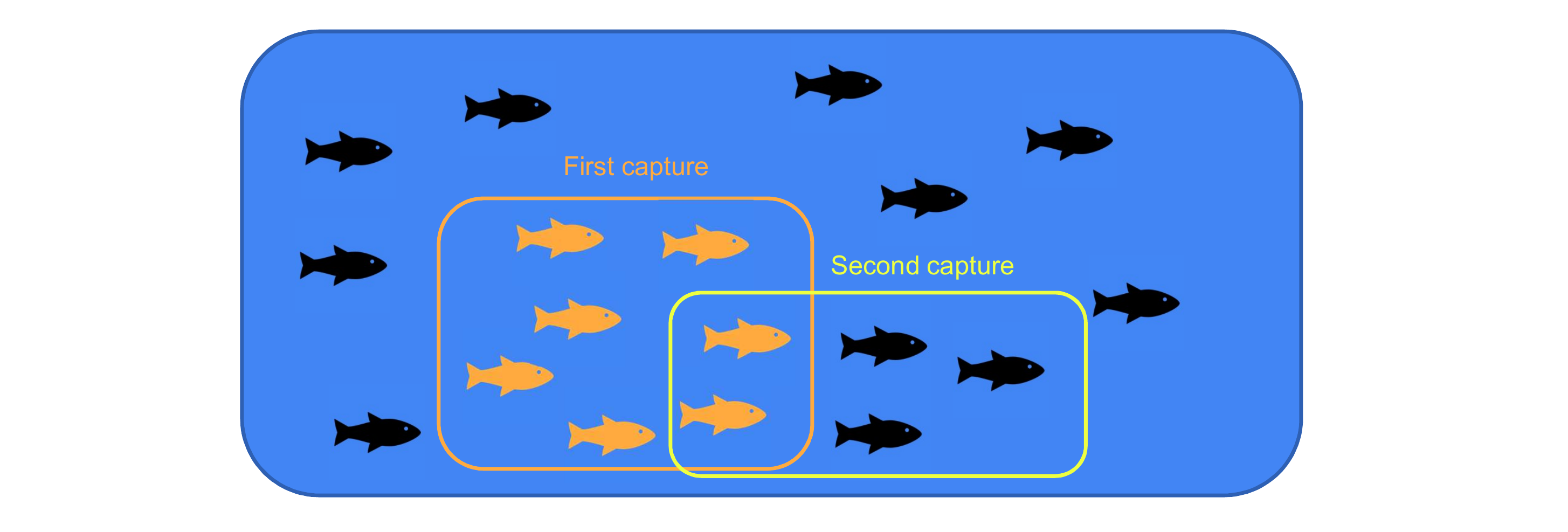}
    \caption{Mark and Recapture: 40\% of fish in the second capture are marked $\Rightarrow$ we marked 40\% of fish in the first capture $\Rightarrow$ the total number of fish is roughly 7/40\% $\approx$ 17. 
}\label{fig:mark-recapture}
\end{figure}

\paragraph{Mark and recapture} is a technique for estimating the total estimation of animals in a certain terrain~\cite{mark-recapture}. For example, assume that a biologist wants to know the number $N$ of fish that live in a given pond. For illustration purposes (Figure~\ref{fig:mark-recapture}), we assume a small number of fish, with $N=18$. The biologist captures a number $M=7$ of fish, and marks them (e.g., with a color mark). Then she releases the fish back into the water (if possible quickly, to avoid that the number of live fish drops to $N-M=11$). The next day, she recaptures a number of fish, and checks how many of them have marks. Let's say that only 40\% of the captured fish is marked. Under the assumption that the ratio $r=40\%$ of marks in the recaptured population is the same as the ratio of marks in the entire population, the biologist can then estimate the total number of fish by the Petersen estimate as $\hat{N}=M/r$ (which is $17$ in our example). This estimator has since been refined by a number of approaches~\cite{mark-recapture}. The question is now how this basic technique can be applied to KBs. It may be tempting to ``capture and release'' entities from the KB, but the technique works only if the entities are captured from the real world, not from the sample that has already been ``captured'' by the KB.

\paragraph{Estimators for collaborative KBs}~\cite{luggen2019non} extend the notion of mark-and-recapture to collaborative KBs such as Wikidata. The main idea is that, by performing an edit on Wikidata, a contributor samples an entity from the real world. Thus, any statement \fact{s}{p}{o} that a user contributes counts as a sample of $s$ and $o$ from the real world. We can then consider all entities that have been edited during a given time period, and that belong to the target class, as one sample. If $D$ is the current number of instances of the target class in the KB, $n$ is the number of entities observed in the sample, and $f_1$ is the number of entities that have been observed exactly once in the sample, the Good-Turing estimator estimates the total number of instances of the target class in the real world as $\hat{N}=D \times (1-f_1/n)^{-1}$. The intuition behind this estimator is as follows: if $f_1=n$, then every entity in the sample has been seen exactly once. This suggests that, as we keep sampling, we will always see new entities. This means that the size of the real-world population is infinite. If, conversely, $f_1=0$, we have seen all entities at least twice during the sampling, which suggests that the class is not bigger than what we currently have ($\hat{N}=D$). Other possible estimators are the Jackknife estimator, an estimator with Singleton Outlier Reduction, and an Abundance-based Coverage Estimator \cite{luggen2019non}.

Experiments on the Wikidata KB show that these estimators converge to the ground truth number of missing entities on 5 of the 8 classes that were tested. Interestingly, some of these classes are \emph{composite classes}, i.e., classes defined by a query such as ``\textit{Paintings by Van Gogh}''.

\paragraph{Estimators for crowd-sourced KBs}~\cite{trushkowsky2013crowdsourced,acosta2017enhancing} target settings where workers are paid to contribute instances to classes or query answers. While every worker has to contribute distinct instances, different workers may contribute the same instances. Workers may also come and go at any moment, and have different work styles. If one worker (a ``streaker'') adds many entities in one go, classical recall estimators may overestimate the total population size (mainly because of a high value of $f_1$, see above). One solution to this problem is to cap the number of new entities per contributor to two standard deviations above the average number of entities per contributor. 

Experiments with the CrowdDB platform\footnote{\url{http://www.crowddb.org/}, down as of September 9, 2022.} 
show that the modified estimator is more conservative in its estimation of the total number of entities. Two experiments with a known ground truth were conducted: For ``UN-recognized countries'', the new estimator converges faster to the true number. For ``U.S. states'', the effect is less visible, most likely because they are easy to enumerate and are thus filled in one go. 

\paragraph{Static estimators} target KBs where entities are not added dynamically. In such KBs, there is no sampling process from the real world. One way to obtain a lower bound for the true population size is to use Benford's Law~\cite{benford1938law}. Benford's Law says that in many real-world sets of numerical data, the 
leading digit tends to be small. More precisely, the 
digit ``1'' appears in roughly 30\% of the cases, the digit ``2'' in 18\% of the cases, and the digit $d$ in $100\times log_{10}(1+d^{-1})$ percent of the cases. This applies in particular to quantities that grow exponentially (such as the population of a city), because a quantity to which a multiplicative factor is applied repeatedly will run through the leading digit ``1'' more often than through the other digits. To apply this technique to KBs~\cite{soulet2018representativeness}, the target class has to have a numerical predicate, e.g., the population size for cities, the length for rivers, the price for products, etc.
This predicate has to obey certain statistical properties for Benford's Law to work. We can then collect the first digits of all numbers, and check whether the distribution conforms to Benford's Law. In the work of \cite{soulet2018representativeness}, this is done primarily to quantify to what degree the sample of entities in the KB is representative of the real world distribution. However, the data can also be used to compute the minimum number of entities that we would have to add to the KB in order for the distribution to conform to Benford's Law. Under the assumption that the real-world distribution follows Benford's Law, this number constitutes a lower bound for the number of missing entities. 

Experiments show that a parameterized version of Benford's Law applies to a number of very diverse predicates, including the population of places, the elevation and area of places, the length and discharge of water streams, the number of deaths and injured people for catastrophes, and the out-degree of Wikipedia pages.

\subsection{Recall of Statements}\label{sec:recall-facts}
After having discussed the recall of entities, let us now turn to the completeness of statements. More precisely:

\problem{Missing Object Problem}{A knowledge base $K$, a subject $s$, and a predicate $p$}{Determine if there is $o$ with \mfact{s}{p}{o}$\in K^i$ 
and \mfact{s}{p}{o}$\not\in K$.}

\noindent
As an example, consider \textit{Marie Curie} and the predicate \emph{wonAward}.  
We are interested in finding whether she won more awards than those given in the KB. It is not always easy to define what we consider missing objects~\cite{what-do-we-actually-know}: does an award from her high-school count? is a public recognition tantamount to an award? is a rejected award still an award she won? etc. In what follows, we assume that we can determine whether a relationship holds, e.g., by a vote from crowd-workers.

\begin{figure}
    \includegraphics[width=0.9\textwidth]{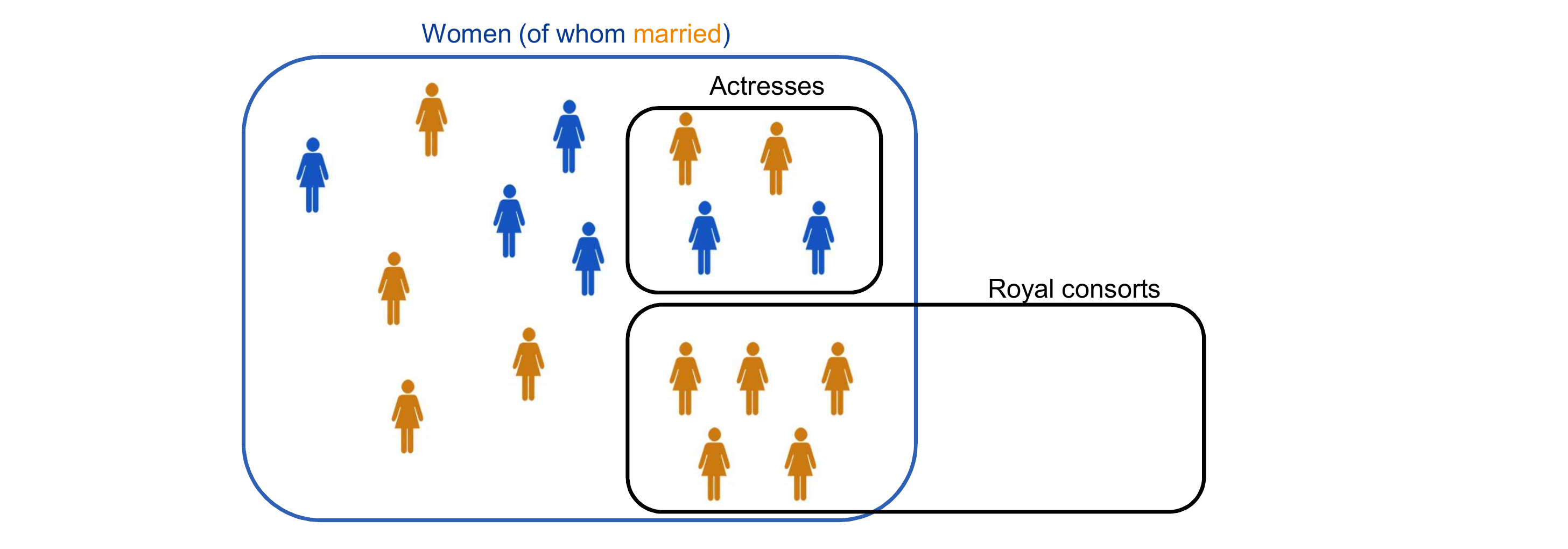}
    \caption{To determine whether being married is an obligatory attribute of the class \emph{Woman}, we check whether the its distribution in the KB changes when we walk into subclasses. 
    }
\label{fig:obligatory}
\end{figure}

A first hint on missing objects comes from \textbf{obligatory attributes}. An obligatory attribute of a given class is a predicate that all instances of the class must have in the real world at the current point in time. For example, \emph{birth\-Date} is an obligatory attribute for humans, while \emph{hasSpouse} is not. If we knew the obligatory attributes, we could use them to solve parts of the Missing Object Problem: if an instance of a class has no object for an obligatory attribute, then it is necessarily incomplete on that attribute. It turns out that the obligatory attributes of a class can be determined automatically from the incomplete KB~\cite{lajus2018all}. For this purpose, we look at the ratio of instances of the class that have the predicate in the KB. For example, we can look at the ratio of women who are married (Fig.~\ref{fig:obligatory}). We then check if this ratio changes when we go into subclasses of the target class, or into intersections of the target class with other classes. In our example, we can go into the subclass of \textit{Actresses}, or into the intersecting class of \textit{Royal Consorts}. If the ratio changes, then a theorem 
tells us (under a number of assumptions) that the predicate \textit{cannot} be obligatory. In our example, the ratio of married women changes when we go into the subclass of \textit{Royal Consorts}. Hence, \emph{married} cannot be an obligatory attribute. If we assume that the other predicates are all obligatory, we can then spot places in the KB where objects must be missing. 

Experiments on YAGO and Wikidata show that the approach can achieve a precision of 80\% at a recall of 40\% over all predicates. The approach can also determine that, for people born before a certain time point, a \emph{death\-Date} becomes an ``obligatory'' attribute.

\paragraph{Weak signals} can be combined to estimate whether a given subject and a given predicate are complete in a given KB~\cite{galarraga2017predicting}. The simplest of these signals is the \emph{Partial Completeness Assumption} 
(PCA), which says that if the subject has some objects with the predicate, then no other objects are missing. Other signals can come from the dynamics of the KB: the \emph{No-Change Assumption} says that if the number of objects has not changed over recent editions of the KB, then it has ``converged'' to the true number, and no more objects are missing. The \emph{Popularity Assumption} says that if the subject is popular (i.e., has a Wikipedia page that is longer than a given threshold), then no objects are missing. A more complex signal is the \emph{Class Pattern Oracle}. It assumes that if the subject is an instance of a certain class $c$, then there are no missing objects. For example, if the subject is an instance of the class \emph{Living\-People}, then no death date is missing. The \emph{Star Pattern Oracle} does the same for predicates: If, e.g., a person has a death place then the person should have a death date.

These signals can be combined as follows: We first add the simple signals as statements to the KB. For example, if \textit{Elvis Presley} is known to be popular, we add \fact{Elvis\-Presley}{is}{popular} to the KB. If Elvis has won several awards, we add \fact{Elvis\-Presley}{moreThan$_k$}{won\-Award} for small $k$. Then, we add the ground truth for some of the subjects. For example, if we know that \textit{Elvis Presley} has won no more awards than those in the KB, we add \fact{Elvis\-Presley}{complete}{won\-Award} to the KB. Finally, we can use a rule-mining system (such as AMIE~\cite{amie}) to mine rules that predict this ground truth. Such a system can find, e.g., that popular people are usually complete on their awards, as in $\fact{s}{\textit{is}}{\textit{popular}} \Rightarrow \fact{s}{\textit{complete}}{\textit{award}}$. These rules can then be used to predict completeness for other subjects, i.e., to predict whether Marie Curie has won more awards than those mentioned in the KB.

Experiments with a crowd-sourced ground truth on YAGO and Wikidata show that some predicates are trivial to predict. These are obligatory attributes with only one or few objects per subject: \emph{bornIn}, \emph{hasNationality}, \emph{gender}, etc. Here, the CWA and PCA work great. For the other relationships, the Popularity Assumption has a high precision, but low recall, i.e., misses that many more subjects than predicted are complete. The Star- and Class-Oracles also do well. The combined approach can achieve F1-values of 90\%-100\% for all 10 predicates that were tested, with the exception of \emph{has\-Spouse}: it remains hard to predict whether someone has more spouses than are in the KB.

\paragraph{Textual information} can also be used to spot incomplete areas of the KB. For example, assume that we encounter the sentence ``Marie Curie brought her daughters Irène and Eve to school''. Then we can conclude that Curie had at least two daughters. In common discourse, we would even assume that she had no other daughters, and in fact no other children at all (if she had other children, we would expect the speaker to convey that, e.g., by saying ``brought only her daughters to school''). This conclusion is an \emph{implicature}, i.e., an information that is conveyed by an utterance (or text) even though it is not literally expressed~\cite{grice1975logic}. In what follows, let us consider a given sentence about a given subject and a given predicate, and let us assume a KB that is complete for that subject and predicate (e.g., Wikidata for popular subjects). If we use a simple open information extraction approach, we can extract the objects that the sentence mentions for the predicate, and compare them to the objects that the complete KB contains~\cite{razniewski2019coverage}. If the sentence mentions all objects, we consider it \emph{complete}. In this way, we can build a training set of complete and incomplete sentences for a given predicate, and we can train a classifier on this set. This classifier can then be used to determine if some other sentence is complete, and if it is, we can determine if the subject of that sentence is complete in the KB. 

Experiments across 5 predicates show that the approach works better on paragraphs than on sentences, because in many cases, the objects are enumerated across several sentences. The F1-values are 45\%-75\%. 

\subsection{Summary}

The recall of a KB can be estimated for entities and for statements. Different approaches have been developed for both cases (Table~\ref{tab:pre}), with promising results. However, the approaches for entities suffer from a small ground truth: in only few cases the total number of instances of a class is known. This makes the experiments difficult to judge.

\begin{table}[]
    \centering
    \begin{adjustbox}{width=\textwidth,center}
    \begin{tabular}{lll}
    \toprule
    Approach & Target & Assumptions\\
    \midrule
         Mark \& Recapture~\cite{mark-recapture}&Entities& Sampling from the real world. \\
         Collaborative KB Estimator~\cite{luggen2019non} & Entities of a given class & Each fact is a random draw from the real world\\
         Crowd-sourced KB Estimator~\cite{trushkowsky2013crowdsourced} & Entities of a given class & Crowd-workers contribute entities independently\\
         Static Estimator~\cite{benford1938law}& Entities of a given class & Distribution follows Benford's Law\\
         Obligatory Attribute Estimator~\cite{lajus2018all}& Relations that are obligatory in the real world & Facts are sampled i.i.d. from the real world\\
         Weak Signal Estimator~\cite{galarraga2017predicting} & Missing objects for a given subject and relation & Sufficient number and quality of weak signals\\
         Extraction from Text~\cite{razniewski2019coverage} & Number of objects for a given subject and relation & Explicit mentions of objects or numbers in text\\
         \bottomrule
    \end{tabular}
    \end{adjustbox}
    \caption{Overview of predictive recall assessment approaches.}
    \label{tab:pre}
\end{table}

\begin{figure}
    \centering
    \includegraphics[width=0.6\textwidth, trim=2cm 3cm 2cm 0cm]{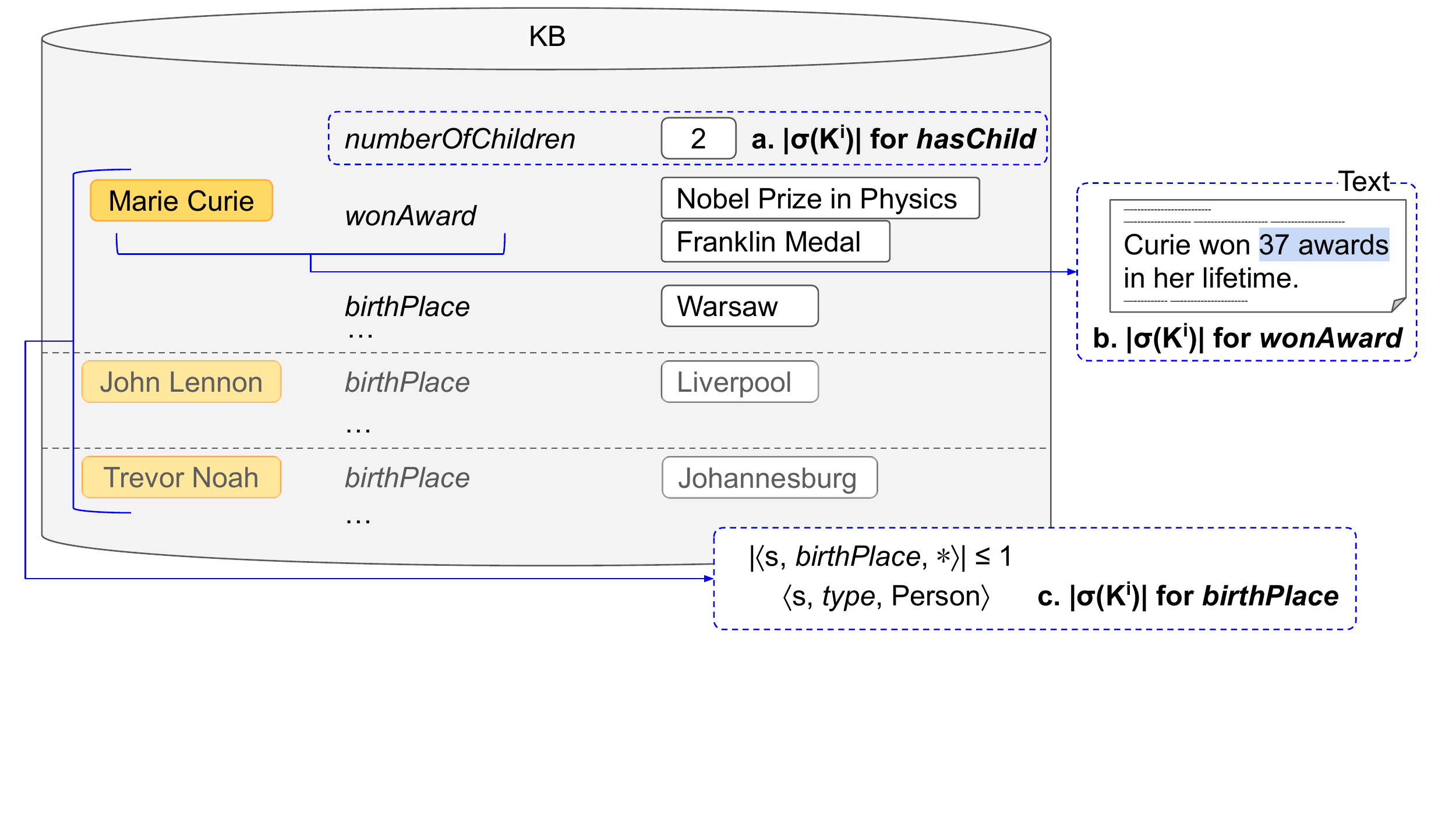}
    \caption{Illustrating three ways of obtaining cardinality assertions (a) from KB predicates (for the number of children), (b) from text (for the number of awards), and, (c) from inference over KB statements (for the number of birthplaces).
    } 
    \label{fig:cardinality_overview}
\end{figure}

\section{Cardinalities from KBs and Text}
\label{sec:cardinalities}
We move from counting entities and objects to identifying cardinality assertions. As defined in Section~\ref{sec:foundations}, cardinality assertions specify the number of records in the ideal KB $K^i$ that satisfy a certain property. But $K^i$ is a hypothetically ideal KB, making the computation of cardinality assertions through selection conditions non-trivial.  
%

Cardinality assertions are expressed explicitly through statements in KB, \fact{Marie Curie}{numberOfChildren}{2}, and in text, \textit{``Curie \underline{won 37 awards} in her lifetime''}. Correctly identified, such statements allow for a direct numeric evaluation of KB recall.
Alternatively, if we know that a certain predicate is complete for a certain subject, we can use that knowledge to deduce the cardinality assertion. 
In this section, we tackle the problem of identifying cardinality assertions from the KB and text, illustrated in Figure~\ref{fig:cardinality_overview}. We look into methods that identify predicates that store cardinality assertions explicitly, for instance, \fact{s}{numberOfChildren}{n}, and statements that ground the cardinality into objects, such as, \fact{s}{hasChild}{$o_i$}, $i \in [0,n]$. 
From now on, we call the predicates that store cardinality explicitly \textit{counting predicates}, $p_c$, and the predicates that store cardinality grounding objects \textit{enumerating predicates}, $p_e$. 


\subsection{Cardinality Information in KBs}
\label{subsec:cardinalityinkb}

The most obvious way to obtain cardinality assertions from KBs would be to use aggregate queries. For example, to determine the number of children \textit{Marie Curie} had, we can count the number of statements satisfying the selection condition, $\mfact{\textit{Marie Curie}}{\textit{wonAward}}{*}$,
but, as we have seen in the previous section, 
KBs suffer from the Missing Objects Problem.
Fortunately, some KBs express cardinality assertions explicitly through counting predicates.
For example, we find the predicates \textit{numberOfGoals} (for the goals scored by football players) and \textit{totalHurricanes} (for the number of hurricanes in a cyclone) in DBpedia and \textit{numberOfEpisodes} (for a film/TV/radio series) in Wikidata as well as in YAGO.
Unfortunately, the predicates of these cardinality assertions usually do not follow any naming scheme: some predicates have the prefix \textit{numberOf...}, but others are specific to the class at hand, including \textit{staffSize}, \textit{population}, or \textit{member count}. Furthermore,
KBs are normally unaware of the semantic relation between the enumerating predicate \textit{hasChild} and the counting predicate \textit{numberOfChildren}, and they do not coexist for all entities. 


Usually, a counting predicate (such as \emph{numberOfChildren}) links a subject to the number of objects of the corresponding enumerating predicate (\emph{hasChild} in this case). However, a counting predicate may also concern the number of \emph{subjects} of some enumerating predicate. For example, the enumerating predicate \emph{worksAt} links a person to their workplace, but the counting predicate \textit{numberOfEmployees} links the workplace to the number of people who work there. To deal with such cases, we assume that the KB contains, for each predicate $p$, also its inverse $p^{-1}$, with all triples. In our example, we assume that the KB contains also the predicate \emph{worksAt}$^{-1}$ (which we call \emph{hasEmployee}). Then \textit{numberOfEmployees} is the counting predicate of the enumerating predicate \textit{hasEmployee}.

Our first task here is to identify these two classes of predicates. Our second task is then to identify a mapping between the two sets of predicates.

\problem{Cardinality Detection Problem}{A knowledge base $K$}{Determine the counting predicates $p_c$ and the enumerating predicates $p_e$. Also determine which $p_e$ corresponds to which $p_c$.}

\paragraph{Cardinality bounding.} 
A first step for cardinality detection can be to bound the cardinality of the predicates. For example, if we know that there are 200 countries in our knowledge base, we know that a person cannot have more than 200 citizenships. This knowledge can then help us find counting predicates.
In some cases, cardinality bounds come from the KB itself. For instance, Wikidata uses the \textit{single-value constraint}\footnote{Single-valued constraint definition in Wikidata: \url{https://www.wikidata.org/wiki/Help:Property_constraints_portal/Single_value}.} for predicates with exactly one object\footnote{An element can have only one atomic number \url{https://www.wikidata.org/wiki/Property:P1086}.}. However, such cases are rare.
%

If the KB can also contain incorrect entries (more realistic than the naive correctness assumption that was brought forward in Chapter 2), one approach to automatically bound cardinalities is by mining significant maximum cardinalities of a predicate for a given class~\cite{giacometti2019mining}.
For a given predicate, \textit{parent}, if we see that a significant portion of the entities in the class \textit{Person} have two \textit{parent} objects, then we can say that the predicate has a maximum cardinality of 2. This cascades to the subclasses of \textit{Person}, for instance to scientists, physicists and so on unless any of the subclass has a tighter bound for significant number of its members. This is closely related to outlier detection, but with a focus on mining generic constraints, with using them for outlier detection being only one of several possible applications. For example, with the above constraint in mind, entities of type person with 6 parents (which are relatively few) could be ignored. 
Using Hoeffding's inequality constraint~\cite{hoeffding1994probability}, a significant maximum cardinality of a class can be reliably mined for a given confidence level and a minimum likelihood threshold. The approach works top down, starting from the class going deeper into its subclasses and pruning very specific constraints for which Hoeffding's inequality would no longer hold true.
This approach works well for finding maximum cardinalities for functional predicates, for instance all humans have one birth year or all football matches have two teams, since these are more stable across entities of a class: majority of persons have one or two parents and very few have more than two. We will see how to tackle entity-variant cardinalities such as books by an author, or destinations of an airline, which can vary between entities, even those belonging to the same class in the next approach. 

\paragraph{Cardinality predicate detection} aims at identifying existing cardinality assertions and the corresponding enumerating predicates in the KB.
A naive approach would be to identify all predicates with positive integers objects as \textit{counting predicates}, and all predicates with one or more KB entities as \textit{enumerating predicates}. 
This naive approach fails due to a number of reasons. First, a positive integer value is a necessary but not sufficient condition for counting predicates. For instance, KB predicates that store identifiers (\textit{episode number}, \textit{VIAF ID} in Wikidata), measurement quantities (\textit{riverLength}) or counts of non-entities (\textit{floorCount}) are not counting predicates. Second, functional predicates such as \textit{birthPlace}, \textit{atomicNumber},
 etc. do not commonly occur with counts. While these predicates do enumerate fixed objects, they do not have any meaningful counting predicate: the fact that the number of birth places of \textit{Marie Curie} is one is not informative. 
Third, many quasi-functional predicates are predominantly functional but also take multiple values, for instance, citizenship is primarily single-valued with many famous exceptions\footnote{Scientists with more than one citizenship during their whole life \url{https://w.wiki/5UR3}}. 

Signals such as the predicate domain, the predicate range, and textual information can be utilized to identify the cardinality predicates~\cite{ghosh2020uncovering}. 
Predicate names provide some clues as to whether it might be an enumerating predicate. For instance, the word \emph{award} in its plural form occurs almost as frequently, if not more than the word in its singular form, but this is not true for birth place (birth places). Objects of enumerating predicates are also entities and encoding the type information of the subjects and the objects of a predicate can also provide clues: a type \textit{Person} (\textit{Curie}) usually co-occurs with multiple instances of the type \textit{Award} (\textit{Nobel Prize in Physics}, \textit{Franklin Medal}). Range statistics such as the mean and the percentile values are also important clues: in DBpedia the number of statements per subject for the predicate \textit{wonAward} is 2.8 on average, and the mean and the 10th percentile values of the predicate \textit{doctoral students} is 28.3 and 5.5, respectively. 



\paragraph{Predicate alignment} aims to align enumerating predicates with corresponding counting predicates. This would allow us to match cardinality assertions with the statements grounding the cardinality.
Aligned predicates can be used to estimate the KB recall, since counting predicates define expected (or ideal) counts for enumerating predicates. If we know that \textit{award} aligns with \textit{numberOfAwards}, then for all entities with \textit{numberOfAwards}, we can compute the recall of their \textit{award} statements. 
In reality, exact matches such as the one above exist only for very few entities and predicates. More often than not the enumerations are incomplete and overlapping. For instance, an institute may use \textit{numberOfStaff} or \textit{numberOfEmployees} to mean the same thing and the corresponding enumerations could come from $\textit{workInstitution}^{-1}$ and $\textit{employedBy}^{-1}$.
Heuristics such as exact and approximate co-occurrence metrics, and linguistic similarity of the predicate labels can be used to suitably aligned pairs~\cite{ghosh2020uncovering}. The number of alignments in a KB thus obtained is much lower than the number of counting and enumerating predicates. For instance, there is no counting predicate that aligns with the enumerating predicate \textit{wonAward}. 
These alignments are directional as well.
For instance, if the score of (\textit{workInstitution$^{-1}$}, \textit{academicStaff}) is greater than that of (\textit{academicStaff}, \textit{workInstitution$^{-1}$}), then an entity with the enumerating predicate \textit{workInstitution$^{-1}$} is more likely to have the counting predicate \textit{academicStaff} than the other way around.

\subsection{Cardinality Information from Text}
\label{subsec:cardinalityintext}
So far we have looked into KBs for cardinality detection, but the KBs are usually sparse. Even if we have identified that \textit{numberOfDoctoralStudents} is a counting predicate, we cannot predict its value for an entity that has the enumerating predicate \textit{doctoralStudent} but no counting predicate \textit{numberOfDoctoralStudents}.
In such cases, we can turn to textual data, i.e., we can tap into textual sources for retrieving cardinality information. For example, if a text says \textit{``Marie Curie advised 7 students''}, we want to extract the number 7 for the predicate \emph{advised}. More formally:

\problem{Counting Quantifier Extraction Problem}{A subject $s$, a text about $s$ and a predicate $p$}{Determine the number of objects with which $s$ stands in predicate $p$ from the text.}

\noindent
The first challenge {we} face is the \emph{variability of text}: the same number can be expressed either as a numeral (``3''), as a word (``three''), or as an expression (``trilogy of books''). The second challenge is \emph{compositionality}: quantities may have to be added up, as in \textit{" ... authored three books and 20 articles"}, which indicates that the predicate \textit{numberOfPublications} has to have the value \textit{3+20=23}. Finally, we have to \emph{determine the target predicate}: The sentences \emph{``advised 7 students''} and \emph{``supervised 7 students''} both map to the target enumerating predicate \emph{advised}.
Several approaches can be used to overcome these challenges.



\paragraph{Open IE methods} \cite{Mausam2012OpenLL,Corro2013ClausIECO} aim to extract relational statements from text, and these can also include cardinality information. For example, from the sentence \emph{Marie Curie has two children}, these methods can extract \fact{Marie Curie}{has}{two children}.
However, this information is not linked to the KB predicate \textit{hasChild}, and it is also not trivial to recover the number 2. 
Roy et al.~\cite{Roy2015ReasoningAQ} take this forward by proposing an IE system that targets quantities. 
It extracts standardized quantity statements of the form $\langle \textit{value}, \textit{units}, \textit{change}\rangle$. This approach works well, and can in principle be applied to cardinality assertions. 
The never-ending learning paradigm (NELL)~\cite{nell} has seed predicates that capture predicate cardinality, such as the \textit{numberOfInjuredCasualties}, but since these are tied to specific cardinality values, it does not learn them for future extractions.

Current state-of-the-art NLP pipelines have made cardinality detection in texts much easier\footnote{https://spacy.io/usage/linguistic-features}. Even though Open IE extractions contain cardinality assertions, extracting the correct count value and mapping to KB predicates is an open challenge. Next we discuss several approaches 
that target specifically the extraction of cardinalities.

\paragraph{Manually designed patterns} can be 
used to extract cardinality assertions from text~\cite{Mirza2016ExpandingWP}.
The idea is to capture compositionality and variability of cardinal information in text which is outside the capability of current Open IE methods. 
\cite{Mirza2016ExpandingWP} proposes 30 manually-curated regular expression patterns for the \textit{hasChild} predicate in Wikidata the extractions achieve more than 90\% precision when compared with a small set of manually-labelled gold dataset and a larger KB available silver dataset. Cardinality assertions were extracted for 86.2k humans in Wikidata of which only 2.65\% had complete \textit{hasChild} statements.    

\paragraph{Automatic extraction of cardinalities} can be modelled as a sequence labelling task where each predicate has its own model. The work of \cite{mirza2018enriching} explores one feature-based and one neural-based conditional random field model. The cardinal information in the input sentence is replaced with placeholders, namely \textit{cardinal}, \textit{ordinal} and \textit{numeric term} so that the model does not learn specific tokens but the concept. The models then input the features such as n-grams and lemmas in the context-window of the placeholder to the feature-based CRF model and in case of a bi-directional LSTM model the input comprises the words, placeholders and character embeddings of the tokens. The model outputs the labels \textit{COUNT} for cardinals, \textit{COMP} for composition tokens and \textit{O} for other tokens, with a confidence score. A sentence can have multiple cardinal labels or multiple sentences can have cardinal information for a given $s$ and $p$. Hence, a consolidation step aggregates cardinals around compositional cues. A cardinality assertion can occur multiple times in a text in which case a heuristic ordering is applied such that cardinal $>$ numeric term $>$ ordinal $>$ article, breaking ties with confidence scores. This follows the intuition that if a text has a sentence \textit{``Curie published 15 articles.''} and another sentence \textit{``Curie published her first article in 1890.''}, then 15 would be the predicted cardinality for Curie's publications.
Existing cardinal predicates in the KB can be selectively used as ground truth
to evaluate the precision of the automatically extracted cardinalities. For example, one can restrict the KB to popular entities, and one can exclude cardinalities 
above the 90th percentile value, as these are most likely outliers. One must be selective,
since incomplete KB information can negatively impact overall training and evaluation. 

\begin{revision}
Given that LLMs have transformed a range of tasks, an obvious question is whether they can be useful for the extraction of cardinality assertions. This broad question comes in two variants: (i) Can LLMs help to extract cardinality assertions from a given text? (ii) Based on the text they have seen during pre-training, can LLMs directly output cardinality assertions? 

The answer to (i) appears a confident yes, as several studies have shown that LLMs can be used as powerful components in textual information extraction \cite{alt2019fine,chu2021knowfi,josifoski2022genie}. Yet such a usage still requires comprehensive efforts around data selection and preparation, information retrieval, and output consolidation, and thus, does not fundamentally transform the task.

A fundamental transformation could be achieved in the second paradigm. As LLMs have seen huge amounts of text during pre-training, one can try to extract relations directly from these models, without further textual input at the extraction stage. 
Although early works have found mixed results~\cite{lin2020birds}, current LLMs like GPT-4 can quite reliably assert counts in the commonsense domain, e.g., counts of body parts (feet, wings, ...) of different animals. For extracting counts of named entities, similar observations hold: As long as the counts are frequently asserted in text, LLMs can return them. However, once the LLMs would need to aggregate themselves, they fall back to common numbers in the relation of interest \cite{singhania2023extracting}. Although LLMs are becoming incrementally more powerful on benchmarks, it appears that Transformer-based architectures exhibit principled limitations, that make correctly solving count tasks over text difficult \cite{helwe2021reasoning}.
\end{revision}



\begin{table}[t]
    \centering
    \begin{adjustbox}{width=0.92\textwidth,center}
    \begin{tabular}{p{1.15cm}p{3cm}
    p{5cm}
    p{6cm}p{6cm}}
    \hline
    \textbf{Source} & \textbf{Work} 
    &  \textbf{Focus} 
    & \textbf{Strengths} & \textbf{Limitations}\\
    \toprule
    KBs  
    & Giacometti et al.~\cite{giacometti2019mining}  
    & Extract maximum cardinalities for class and predicate pairs.
    & Provides significance guarantees. \newline
    Scalable for large KBs. \newline
    Efficient pruning to reduce search space. 
    & Majority predicates identified have maximum cardinality of 1. \newline 
    Scope for discovering non-functional assertions is limited.\\
     \\
    & Ghosh et al.~\cite{ghosh2020uncovering} 
    & Extract counting and enumerating predicates.
    & Provides important features for identifying cardinality predicates. \newline
    Maps counting predicates to matching enumerating predicates. 
    & Statistical cues provide weak signals. \newline
    Textual cues have limited informativeness.
    \\
    \midrule
    
    Open IE 
    & Mausam et al.~\cite{Mausam2012OpenLL} \newline Corro et al.~\cite{Corro2013ClausIECO}  
    & Extract all \fact{s}{p}{o} statements.
    & Captures cardinality information present as counts. 
    & Canonicalization required to incorporate statements in to KBs. \\

    \\
    &  Carlson et al.~\cite{nell} 
    &  Extract new statements and rules.
    &  Never-ending learning paradigm (NELL). \newline
    Infers new predicates.
    & Cannot learn cardinality assertions from seed predicates. \\
    \\
    
    & Roy et al.~\cite{Roy2015ReasoningAQ} 
    & Extract quantity statements.
    & In principle can extract cardinality assertions.
    & Has only been used for quantity statements. \\

    \midrule
    
    Text
    & Mirza et al.~\cite{Mirza2016ExpandingWP}  
    & Compute cardinality assertions.
    & Manual patterns are effective in cardinality extractions.
    & Scaling predicate-specific patterns. \\
    \\
    
    & Mirza et al.~\cite{mirza2018enriching}  
    & 
    & Sequence modelling for automatic extraction of cardinality information. 
    \newline Consolidation across compositions and multiple mentions.  
    & Prior knowledge of cardinality predicates.
    \newline Zero counts extraction is limited to ad-hoc preprocessing. \\
 
    \hline
    \end{tabular}
    \end{adjustbox}
    \caption{Comparison of related works on cardinality information.}
    \label{tab:cardinality_information}
    
\end{table}

\subsection{Summary}
In this section we have defined the task of identifying cardinalities for predicates. We have seen how to extract cardinalities from KBs and from textual data. Cardinality information has many applications and yet unsolved challenges. We have compiled the related works covered in this section, their focus, strengths and limitations in Table~\ref{tab:cardinality_information}.


\paragraph{Applications.} The CounQER system
\cite{ghosh2020counqer} demonstrates the usefulness of \textbf{aligned predicates} in a simple QA setting 
where given a subject $s$ and (an enumerating) a counting predicate $p$ the system 
returns the objects that satisfy \fact{s}{p}{*} and statements from top-5 aligned (counting) enumerating predicates if available. For instance, in DBpedia, if we look for the \textit{Royal Swedish Academy of Sciences}, the enumerating predicate \textit{workplaces$^{-1}$} returns eight entities who work there, but {we} also learn that the aligned counting predicates \textit{academicStaff} and \textit{administrativeStaff} are unpopulated. Again in DBpedia, {we} learn that the enumerating predicate \textit{doctoralStudent} has no corresponding counting predicate: For the subject \textit{Marie Curie} and the predicate \textit{doctoralStudent}, the system returns 7 entities who were her doctoral students but no aligned cardinality assertions. 

Tanon et al.~\cite{pellissier2017completeness} use cardinality information to improve rule-mining. Rule mining is the task of finding interesting associations between entities in a KB, such as: if \fact{?x}{hasSpouse}{?y} and \fact{?y}{livesIn}{?z} then \fact{?x}{livesIn}{?z}. Such rules can then be used to predict the place of residence of someone.
Cardinality information can avoid that we predict too many such places of residence per person, by downgrading the scores of predictions that violate (soft) cardinality constraints.
The authors evaluate the recall-aware scores against standard scoring metrics and find that recall-aware scores highly correlate with rule quality scores in the setting of increasingly complete KBs. 
Another line of work uses cardinality information as priors in neural link prediction~\cite{Muoz2019EmbeddingCC}. Similar to the work by Tanon et al.~\cite{pellissier2017completeness}, they regularize the number of high probability predictions by penalizing the model when the number of predictions violate the cardinality bounds of a given relation type.  
\paragraph{Open challenges.} 
\begin{revision}
LLMs could in principle be used to infer commonsense cardinality information such has the number of parents a person has or specific cardinality assertions, such as Marie Curie advised 7 doctoral students. 
Experiments in probing older LLMs for numerical commonsense ~\cite{lin2020birds} show that fine-tuning improves the model performance, though the models could not surpass humans. 
The 2023 edition of the LM-KBC challenge~\cite{lm-kbc-challenge-2023}, whose main focus is on constructing KB from an LLM, contains cardinality prediction for two relations, \textit{numberOfChildren} of a person and \textit{numberOfEpisodes} of a TV series. Even the best-performing system, that relied on GPT-4, achieved only 69\% F1-score on both relations. It appears that more work is needed here before LLM outputs could reliably feed a knowledge base. 
A common challenge to both tasks of extracting cardinality assertions and aligning counting with enumerating predicates  is generating high-quality training and ground truth data. There is no single authoritative source on groundtruths: We have IMDB for movies\footnote{IMDB: \url{https://www.imdb.com/}}, cast and crew, and the GeoNames dataset for geographical locations\footnote{GeoNames\url{https://www.geonames.org/}}. These are great examples of high recall datasets, but the situation is not as rosy when we move to other topics, like scientists or monuments. Crowd-sourcing is an option, but it poses several challenges, most importantly data quality and scalability. The other option commonly used is to employ heuristics. As we saw for cardinality extraction from text~\cite{mirza2018enriching}, distant supervision can be used to extract ground truth statements, with certain restrictions, such as relying on popular entities and KB statistics, such as the $90^{\textrm{th}}$ percentile predicate value to filter out possible outliers.    
\end{revision}


\section{Identifying Salient Negations}
\label{sec:negation}

Knowledge bases store by and large positive knowledge, and very little to no negative statements.
\begin{revision}
This happens for principled reasons, as the set of possible negations is vast and possibly infinite (depending on whether one assumes a finite or infinite set of constants). Adding complete sets of negative statements is therefore hardly a goal.  Many standard AI applications, such as question answering and dialogue systems, would often benefit from statements about popular entities, i.e., explicit negations for salient cases. For instance, a cooking chatbot should be aware that certain ethnic food are not meant to be heated, e.g., \textit{Hummus} and \textit{Gazpacho}, and a general-purpose search engine must be confident about common factual mistakes, such as famous people not winning certain awards in their domains, e.g. \textit{Stephen Hawking} and \textit{The Nobel Prize in Physics}. Completeness statements enable the inference of negations, yet are themselves hard to come by. Furthermore, even though negative statements are in principle much more numerous than positive statements, only few of them are interesting.
In this section, we approach the problem of negation materialization therefore not with the goal of completeness, but with the goal of a \textit{high recall among salient negations}. 

The concept of \textit{salience} has a long history in psychology \cite{taylor1978salience} as well as in sociolinguistics \cite{racz2013salience}, and there are also recent attempts to capture it in knowledge bases \cite{klein2022identifying}. Approaches to modelling salience typically revolve around the concepts of frequency, unexpectedness, or self-interest (for acting agents), yet universal agreement is lacking. Furthermore, none of the models is easy to operationalize. The works that we present next therefore usually utilize human (crowd worker) judgements as yardstick for salience. 
\end{revision}

We first review existing negative knowledge in open-world knowledge bases. We then show how \textit{salient negative statements} can be automatically collected from within incomplete KBs, and via text extraction.

\subsection{Negation in existing KBs} 
\label{subsec:existingnegation}
Web-scale KBs operate under the Open-world Assumption (OWA). This means that an absent statement is not false, but only unknown. The only ways to specify negative information is to either explicitly materialize negative statements in the KB, or to assert constraints that implicitly entail negative statements.
In this section, we focus of the former case. Even though most KB construction projects do not actively collect negative statements,
a few of them allow implicit or explicit negative information:
\begin{itemize}
    \item Negated predicates: a few KBs contain predicates that express negative meaning, i.e., contain negation keywords. For example, DBpedia~\cite{dbpedia} has predicates such as \textit{carrierNeverAvailable} for phones, and \textit{neverExceedAltitude} 
    for airplanes. The medical KB Knowlife~\cite{ernst2015knowlife} contains predicates such as \textit{isNotCausedBy} and \textit{isNotHealedBy}. Wikidata allows a few type-agnostic negated predicates, namely \textit{differentFrom} (827000 statements), \textit{doesNotHavePart} (535 statements), \textit{doesNotHaveQuality} (422 statements), \textit{doesNotHaveEffect} (36 statements), and \textit{doesNotHaveCause} (13 statements). A more systematic example for negated predicates can be found in ConceptNet~\cite{conceptnet}, where the 6 main predicates have negated counterparts, namely \textit{NotIsA}, \textit{NotCapableOf}, \textit{NotDesires}, \textit{NotHasA}, \textit{NotHasProperty}, and \textit{NotMadeOf}. Yet the portion of negative knowledge is less than 2\%. Furthermore, many of the negative statements are uninformative, as in \fact{tree}{NotCapableOf}{walk}.
    \item Count predicates: a subtle way to express negative information is by matching count with enumeration predicates 
    (see Section~\ref{subsec:cardinalityinkb}). For example, if a  KB asserts \fact{Marie Curie}{numberOfChildren}{2}, accompanied by two hasChild-statements, this indicates that for this subject-predicate pair, the list of objects is complete. Therefore, no other entity is a child of \textit{Curie}. 
    \item Statements with negative polarity: In the Quasimodo KB~\cite{romero2019commonsense}, every statement is extended by a polarity value to express whether it is a positive or a negative statement, e.g., \fact{scientist}{has}{academic degree} with \textit{polarity=positive}. Quasimodo contains a total of 351K negative statements.
    \item No-value objects: Wikidata~\cite{wikidata} allows the expression of universally negative statements, where a subject-predicate pair has an empty object. For example, \fact{Angela Merkel}{hasChild}{no-value}\footnote{\url{https://www.wikidata.org/wiki/Q567}}. The total number of such statements with a no-value object in Wikidata is 20.6K
    \item Deprecated rank: KBs like Wikidata encourage flagging certain statements as incorrect as opposed to removing them. These are usually outdated statements or statements that are known to be false\footnote{\url{https://www.wikidata.org/wiki/Help:Deprecation}}, with a total of 20.4k statements.
\end{itemize}

\begin{figure}[t]
    \centering
    \includegraphics[width=0.9\columnwidth]{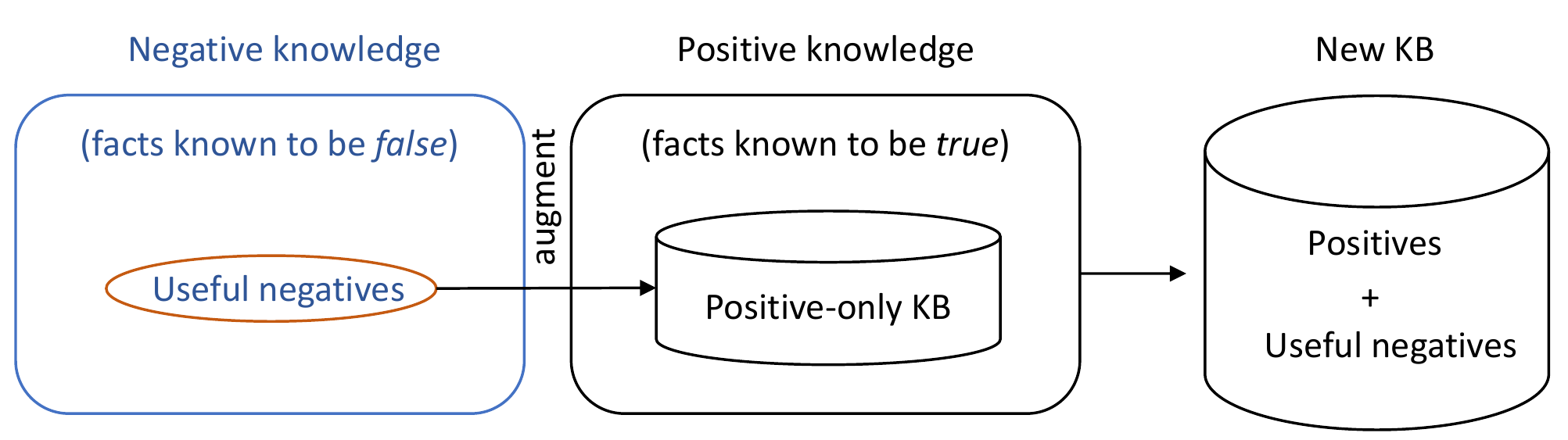}
    \caption{The main challenge is to identify the subset of \textit{useful} negative statements that can be added to an existing positive-only open-world KB.}
    \label{fig:negationproblemoverview}
\end{figure}

While these notions give us a way to express negative statements, one still have to find a way to add them -- if possible, automatically. The key challenge is to focus on identifying \emph{salient} negative statements, such as a well-known physicist not winning the \textit{Nobel Prize}, namely $\neg$\fact{Stephen Hawking}{wonAward}{Nobel Prize in Physics}. An illustration of the research problem is shown in Table~\ref{fig:negationproblemoverview}. The number of false facts or statements is much larger than positive facts, e.g., Stephen Hawking studied at a 4 educational institution v. thousands that he did not study at. The key is to identify the subset of \textit{useful} negative statements that can be added to the existing open-world KB.

\begin{revision}
As discussed above, we take a model-free approach here and leave the choice of what is salient to human annotators. We hypothesize only that frequency and unexpectedness are likely ingredients \cite{taylor1978salience,racz2013salience} of their decision. Where human judgments cannot be obtained, one could resort to an automatically computed target metric like proposed in \cite{klein2022identifying}, though at the risk of optimizing for what can be conveniently computed, instead of what is truly salient. \end{revision}

\problem{Salient Negations Problem}{A subject $s$ in a web-scale incomplete KB }{Identify accurate and salient negative statements about $s$.}

\noindent
We divide approaches into three main categories: methods that use well-canonicalized KBs, methods that use loosely-structured KBs, and methods that use text corpora as the main source of negative statements.

\subsection{Salient Negations in Well-structured KBs}

Famous KBs such as Wikidata, YAGO, and DBpedia consist of well-canonicalized statements (with minimal ambiguity), i.e., the A-box, and are accompanied with manually crafted schemas (aka the T-box). The following approaches discover interesting negations about entities by relying on positive statements from well-structured KBs.

\paragraph{Peer-based inferences}\cite{arnaout020enriching} is one of the earliest approaches to solve this problem. It proposes deriving candidate salient negations about a KB entity from \textit{highly related entities}, i.e., peers, then ranks them using relative statistical frequency. For example, \textit{Stephen Hawking}, the famous physicist, has never won a {Nobel Prize in Physics}, nor has he received an {Academy Award}. However, highly related entities (other physicists who are Nobel Prize winners) suggest that not winning the Nobel Prize is more noteworthy for Hawking than not winning the Academy Award. The first step must thus be to identify the relevant peers for the input entity. The method of~\cite{arnaout020enriching} offers three similarity functions to collect peer entities for a given input entity: (i) class-based similarity~\cite{balaraman2018recoin}: This takes advantage of the type-system of the KB by considering two entities as peers if they share at least one type; (ii) graph-based similarity: This relies on the number of predicate-object pairs that two entities have in common; and (iii) embedding-based similarity: This captures latent similarity between entities by measuring the cosine similarity of their embeddings~\cite{wikipedia2vec}.

\begin{example}
Using (i), \textit{Stephen Hawking} is a physicist like \textit{Max Planck} and \textit{Albert Einstein}, hence, information about \textit{Planck} and \textit{Einstein} can help in creating candidate negations for \textit{Hawking}. Using (ii), \textit{Stephen Hawking} and \textit{Boris Johnson} share 9 predicate-object pairs, including \textit{(gender, male)}, \textit{(citizenship, U.K.)}, and \textit{(native language, English)}. Using (iii), one of the closest entities to \textit{Hawking} using this measure is his daughter \textit{Lucy Hawking}.
\end{example}

\noindent
Negations at this point are only candidates. Due the KB's incompleteness, each candidate could be true in the real world, and  just be missing from the KB. The peer-based method requires the candidates statements to satisfy the partial completeness assumption (the subject has at least one other object for that property~\cite{amie,AMIEP}) in order to be considered for the final set. In particular, if \textit{some} awards are listed for the subject \textit{Stephen Hawking} then the list of awards is assumed to be \textit{complete}, and any missing award is absent due to its \textit{falseness}. 
\begin{example}
We know that \textit{Hawking} has won other awards such as the \textit{Oskar Klein Medal} and has children including \textit{Lucy Hawking}, but we know nothing about his \textit{hobbies}. Therefore, the candidate statement $\neg$\fact{Stephen Hawking}{hasHobby}{reading} is discarded.
\end{example}

\noindent
 In an evaluation over correctness of inferred negations~\cite{arnaout020enriching}, this simple yet powerful rule increases the accuracy of results by 27\%. Remaining candidate negations are finally scored by relative peer frequency. For instance, 2 out of 2 peers of \textit{Hawking} have won the \textit{Nobel Prize in Physics} but only 1 out of 2 is the parent of \textit{Eduard Einstein}.

\paragraph{Order-oriented peer-based inferences}~\cite{ARNAOUT2021100661} is an extension of the previous method where KB qualifiers such as temporal statements are leveraged to obtain better peer entities and provide negations with explanations. In this order-oriented method, one input entity can receive \textit{multiple} peer groups. For instance, 3 separate peer groups for \textit{Max Planck} are \textit{winners of the Nobel Prize in Physics}, \textit{winners of Copley Medal}, and \textit{alumni of Ludwig Maximilian University of Munich}. In addition to peering, ranking also accounts for the proximity between the input and peer entities.

\begin{example}
The negative statement $\neg$\fact{Max Planck}{educatedAt}{The University of Cambridge} with provenance ``unlike \textit{the previous 3 out of 3} winners of the \textit{Nobel Prize in Physics}'' is favored over $\neg$\fact{Max Planck}{citizenOf}{France} with provenance ``unlike \textit{3 out of the previous 18} winners of the \textit{Nobel Prize in Physics}''.
\end{example}
In this example, \textit{temporal recency} is rewarded. In other words, the same peer frequency that is far behind in the ordered peers will receive a lower score. Moreover this work introduces the notion of \textit{conditional negation}, as opposed to the previous simple ones. While a simple negation is expressed using 1 negative statement, $\neg$\fact{Albert Einstein}{educatedAt}{Harvard}, a conditional negation goes beyond 1 to express negative information that is true \textit{only} under certain condition(s).

\begin{example}
$\neg\exists o$ \fact{Albert Einstein}{educatedAt}{$o$}\fact{$o$}{locatedIn}{U.S.} (meaning \textit{Einstein} has never studied at a university in the \textit{U.S.}).
\end{example}

\noindent
In a crowdsourcing task to evaluate the quality of result negations in~\cite{ARNAOUT2021100661} the peer-based method achieves a 81\% in precision and 44\% in salience, while the order-oriented inference improves both the precision and salience, to 91\% and 54\% respectively.

\subsection{Salient negations in Loosely-structured KBs}
\label{subsec:negation:csk}

While encyclopedic KBs like Wikidata, YAGO, and DBpedia are well-canonicalized, commonsense KBs like ConceptNet~\cite{conceptnet} and Ascent~\cite{ascent,ascentpp} express information using uncanonicalized short phrases. For instance, Ascent contains the actions \textit{lay eggs}, \textit{deposit eggs}, and \textit{lie their eggs}. Therefore, the methods designed for well-structured KBs will result in many incorrect inferences, e.g., \fact{butterfly}{capableOf}{lie their eggs} but $\neg$\fact{butterfly}{capableOf}{lay eggs} and $\neg$\fact{butterfly}{capableOf}{deposit eggs}. Moreover, the use of the PCA rule (only infer an absent object to be negative in the presence of at least one other object for the same predicate) to improve accuracy of inferred negations would not be sufficient as most commonsense KBs rely on ConceptNet's well-defined but handful and generic predicates, e.g., \textit{hasProperty}. As opposed to Wikidata's \textit{citizenOf} predicate where only a few objects are expected, ConceptNet's \textit{hasProperty} can have hundreds of accepted object phrases. Hence, it is not a very good idea to assume that since an entity has at least one general property present the list of objects is complete.

\begin{revision}
\paragraph{NegatER}~\cite{safavi-etal-2021-negater,safavi2020generating} is a recent method for identifying salient negations in commonsense KBs. Given a subject $s$ (an everyday concept in this case) and an input KB, a pre-trained language model (LM) is fine-tuned using the training and testing sets of KB statements. The positive statements are simply queried from the KB, while negative samples are randomly generated under the closed-world assumption, i.e., the KB is complete and the negations are generated by corrupting parts of positive statements with any other random concept or phrase. The LM is then trained to learn a decision threshold per predicate, and hence, the fine-tuned LM serves as a true/false classifier for unseen statements. In order to create the set of \textit{informative} or \textit{thematic} negations for $s$, $s$ is replaced by a neighboring \textit{entity phrase} from the KB.

\begin{example}
\fact{horse}{isA}{pet} is replaced by the neighbor subject \textit{horse rider}, resulting in the candidate negation $\neg$\fact{elephant}{isA}{pet}.
\end{example}

\noindent
The classifier (LM) then decides on the \textit{falseness} of such candidate. Once the set of candidates is constructed, they are finally ranked using the fine-tuned LM by descending order of negative likelihood. Even though NegatER compiles lists of thematically-relevant negatives, one major limitation is that it generates many type inconsistent statements, due to the absence of a taxonomy, e.g., $\neg$\fact{horse rider}{isA}{pet}.
\end{revision}

\paragraph{Uncommonsense}~\cite{arnaoutcikm2022} identifies salient negations about target concepts (e.g., \textit{gorilla}) in a KB by computing comparable concepts (\textit{e.g., zebra, lion}) using external structured taxonomies (e.g., WebIsALOD~\cite{webisalod}) and latent similarity (e.g., Wikipedia embeddings~\cite{wikipedia2vec}). Similar to the peer-based negation inference method~\cite{arnaout020enriching}, the PCWA is used to infer candidate negations (e.g, \textit{has no tail}, \textit{is not territorial}). A crucial difference is the technique used for checking the accuracy of candidate negations. As previously mentioned, the PCA would not be sufficient in loosely-structured KBs. Therefore, this work introduces different scrutiny steps to improve the accuracy of candidates. It performs semantic similarity checks (using~\cite{reimers2019sentence}) against the KB itself, and external source checks (using pretrained LMs). Semantic similarity also contributes in grouping negative phrases with the same meaning in order to boost their order in the final ranked list, i.e., the relaxed sibling frequency. The finally generated top-ranked negations are extended with provenances showing why certain negations are interesting. For example, \textit{gorilla} has no tail, unlike other land mammals such as \textit{lions} and \textit{zebras}.
On 200 randomly sampled concepts with their top-2 negations~\cite{arnaoutcikm2022}, in a crowdsourcing task where workers are asked about the accuracy and interestingness of given negative statements, UnCommonSense achieves a precision of 75\% and a salience of 50\%, while NegatER achieves a similar precision of 74\% but lower salience of 29\%.

\subsection{Salient Negations in Text}
\
Large textual corpora can be good external sources for implicit and explicit negations. Moreover, due to the incompleteness of existing KBs, text-based methods can be complementary to inference-based methods.

\paragraph{Mining negations from query logs}~\cite{arnaout020enriching} is an unsupervised pattern-based methodology that extracts salient negations from query logs. The intuition is that users often ask why something does not hold, as in ``\textit{Why didn't Stephen Hawking  win the Nobel Prize in Physics?}''. Such a query can then be used to deduce that \textit{Hawking} did indeed not win the \textit{Nobel Prize}.
The work defines 9 negated why-questions, such as \textit{Why didn't <s>?}. 
 \begin{example}
 Given the pattern \textit{Why didn't Stephen Hawking..}, the auto-completion API of a search engine, e.g., Google's, produces \textit{won the Nobel Prize in Physics, accept the knighthood, ...}.
 \end{example}

\begin{revision}
\paragraph{Mining negations from text}~is also a possibility, but it has not 
led to significant results so far~\cite{hiba-thesis}. The reason is that long texts (such as newspapers, blogs, and encyclopedias) rarely mention salient negations, as an analysis on the STICS~\cite{STICS} corpus has shown. Sentences with a negative meaning in newspapers and blogs 
are mostly about things that people did not do or did not say -- as in ``\textit{Brad Pitt did not threaten Angelina Jolie with cash fine}'', 
or ``\textit{Angela Merkel never made much of an effort to ensure that eastern Germans felt a sense of belonging}''.  Encyclopedias, on the other hand, mostly contain only positive statements. The few sentences that do contain negation usually contain either double negation, temporary negatives (as in ``\textit{Hawking was \underline{not initially} successful academically}''), or negations of specification (as in ``\textit{His family \underline{could not} afford the school fees \underline{without} the financial aid of a scholarship}''). Overall, none of these sources contain short trivia sentences with negative keywords. 
\end{revision}

\paragraph{Mining negations from edit logs}~\cite{karagiannis2019mining} is a work that exploits the edit history of collaborative encyclopedias such as Wikipedia as a rich source of implicit negations.
Editors make thousands of changes everyday for various reasons, including fixing spelling mistakes, rephrasing sentences, updating information on controversial topics, and fixing factual mistakes. The work focuses on mining data from the last category. In particular, it looks at sentence edits in Wikipedia where only 1 entity or 1 number is changed.

\begin{example}
``The discovery of uranium is credited to \textit{Marie Curie}'' is updated to ``The discovery of uranium is credited to \textit{Martin Heinrich Klaproth}'' where the entity \textit{Marie Curie} is replaced with \textit{Martin Heinrich Klaproth}. The former is then considered a common mistake, i.e., an interesting negative statement.
\end{example}
To decide whether an update must be labeled as \textit{common factual mistake} or one of the other categories, a number of heuristics are applied. These include (i) checking how often is sentence is being updated (to exclude controversial topics where different editors have different opinions); (ii) computing the edit distance between the entities (to exclude spelling corrections). 
and (iii) checking for synonyms (to exclude simple rephrasing of the same statement).
It remains to be checked whether the edit removes or introduces a false statement. 
This is done by counting the number of supporting statements 
on the web.

\subsection{Summary}
In this section we defined the task of identifying salient negations about KB subjects and presented different approaches to tackle this problem. A summary of these approaches with their focus, strengths, and limitations is shown in Table~\ref{tab:negation_comparisons}. 

\begin{figure}[t]
    \centering
    \includegraphics[width=0.85\textwidth]{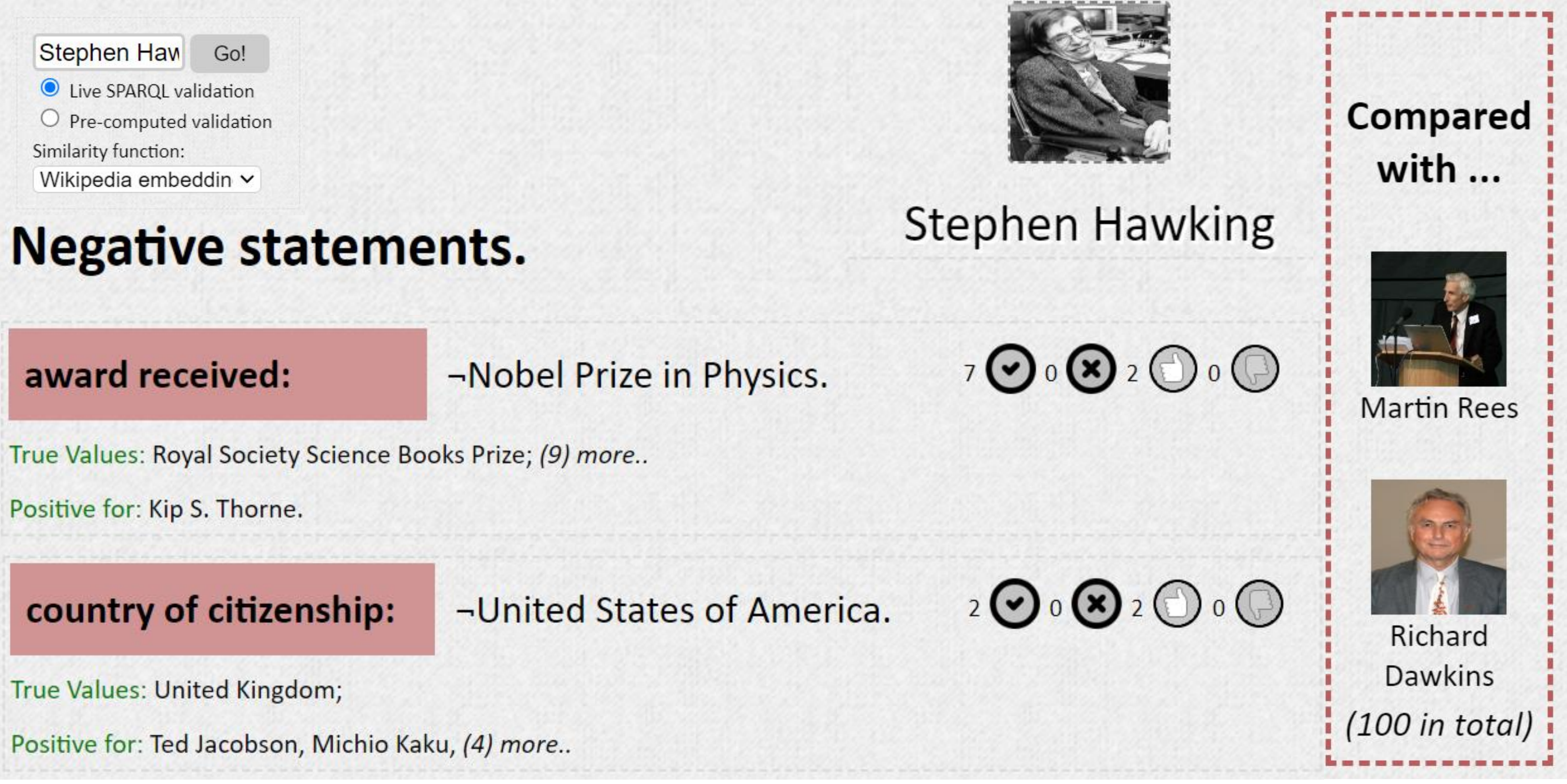}
    \caption{User interface of the WikiNegata (\url{https://d5demos.mpi-inf.mpg.de/negation}) platform. It shows automatically computed salient negative statements for Stephen Hawking, such as that he did not win the Nobel Prize in Physics, unlike his colleague and friend Kip Thorne.}
    \label{fig:wikinegata}
\end{figure}

\paragraph{Applications. } Wikinegata~\cite{arnaout2021wikinegata,arnaout2021negative} is a tool for browsing more than 600m negations about 0.5m Wikidata entities. It gives insights into the peer-based method~\cite{arnaout020enriching} where users can inspect different peers used to infer certain negations (see Figure~\ref{fig:wikinegata}). In commonsense KBs, the Uncommonsense system~\cite{arnaoutcikm2022} provide a similar experience about everyday concepts. 

\paragraph{Open challenges.} For text-based methods, the main issue seems to be the subject recall. Often, negations are only expressed when they are highly exceptional and about prominent entities. For KB-based methods, the main problem is the precision-salience trade-off. It is quite simple to get a near perfect precision when assuming the CWA, as the majority of inferred negations would be correct but nonsensical, e.g., \textit{Stephen Hawking's capital is not Paris}. According to~\cite{arnaoutcikm2022}, this baseline receives a 93\% in precision but less than 7\% in salience. As shown in this section, it becomes challenging to increase salience while maintaining a high level of precision, as plausible candidates tend to be harder to scrutinize, especially in commonsense KBs, e.g., \textit{Is Basketball hard to learn?}. Moreover, negation generators have to deal with real-world changes. This is especially true for the encyclopedic KBs where new information is added frequently. For example, prior to 2016 it was interesting that \textit{Leonardo DiCaprio} never received an Oscar (the negation is no longer correct). A second challenge is the class hierarchy for both entity peer measures and negation generation. For example, noisy taxonomies would result in irrelevant peers.  There are also modeling issues and inconsistencies that most web-scale KBs suffer from, especially the collaborative ones. For instance, to express that a person is vegan, should the editors use \fact{person}{lifestyle}{veganism} or \fact{person}{isA}{vegan}. While one can be asserted, the other could be mistakenly negated by one of the discussed methods. \begin{revision}Finally, the methods presented in this section are meant to compile a list of simple negative statements about commmonsense and encyclopedic entities. Complex negatives, on the other hand, need further investigation and are more challenging. In scientific knowledge, for instance, two contradictory facts might be true under different contexts, e.g. \textit{water cannot extinguish every type of fire}, such as \textit{petrol fires}, but that does not mean that \textit{water cannot extinguish fire}. Also, in socio-cultural knowledge, the same statement can be both true and false under different cultural factors, e.g., \textit{drinking wine at weddings} (in \textit{Europe} v. the \textit{Middle East}).\end{revision}

\begin{revision}
\paragraph{LLMs for Negation Generation.}     Very recent studies examined the ability of large language models (LLMs), such as chatGPT~\cite{chatgpt}, to generate salient negative statements~\cite{hiba-llms-negation,chen-etal-2023-say}.
Findings in~\cite{chen-etal-2023-say} are that contradictions exist in the LLM's belief, when comparing results of different tasks targeting the same piece of knowledge.
For instance, LLMs generate the sentence ``\textit{Lions live in the ocean}'', but answer ``\textit{No}'' when asked ``\textit{Do lions live in the ocean?}''. 
Lessons from~\cite{hiba-llms-negation} include the importance of prompt engineering in this task. Prompts with expressions ``negative statements'', ``negated statements'', and ``negation statements'' return very different types of responses.
Moreover, LLMs struggle with the true negativity of the statements returned, often generating statements with negative keywords but a positive meaning, e.g., ``a coffee table is not only used indoors''. 
\end{revision}


\begin{table}[t]
    \centering
    \begin{adjustbox}{width=\textwidth,center}
    
    \begin{tabular}{p{1.15cm}p{3cm}
    p{5cm}
    p{6cm}p{6cm}}
    \hline
    \textbf{Source} & \textbf{Work} & \textbf{Focus} & \textbf{Strengths} & \textbf{Limitations}\\
    \toprule
      KB & Arnaout et al.~\cite{arnaout020enriching,arnaout2021wikinegata,arnaout2021negative} &  Interesting negations about encyclopedic entities using peer-based statistical inferences. & Subject recall. \newline  \newline  Salience due to peer frequency measures. &  Precision due to KB incompleteness \& modelling issues. \newline \newline Beyond simple negations (conditional negations).\\
     \hline 
    Query logs & Arnaout et al.~\cite{arnaout020enriching}  & Interesting negations about encyclopedic entities using pattern-based query extraction. & Precision due to high-quality search engine query logs. & Subject recall due to APIs access limit. \\
     \hline 
   Edit logs &  Karagiannis et al.~\cite{karagiannis2019mining}  & Common factual mistakes using mined textual change logs. & Precision due to heuristics including web hits computation. & Focus on precision over salience \newline  \newline Mined negations require canonicalization.\\
     \hline 
      KB  & Arnaout et al.~\cite{arnaoutcikm2022} & Informative negations about everyday concepts using fine-tuned LMs. & Salience due to comparable concepts \newline \newline Interpretable results through provenance generation. \newline \newline Can handle non-canonicalized KBs. & Recall depends on presence of subject in the external taxonomy.\\
    \hline
     KB and LM & Safavi et al.~\cite{safavi-etal-2021-negater,safavi2020generating}   & Informative negations about everyday concepts using comparable taxonomic siblings. & Recall through corruptions using phrase embeddings. & Plausibility due to taxonomy not being considered.\\
   \hline
    \end{tabular}
    \end{adjustbox}
    \caption{Comparison of different works on salient negation in KBs.}
    \label{tab:negation_comparisons}
    
\end{table}

\section{Relative Recall}
\label{sec:relativerecall}
\begin{figure}[t]
    \centering
    \includegraphics[width=0.9\columnwidth,trim=2cm 3cm 2cm 0cm]{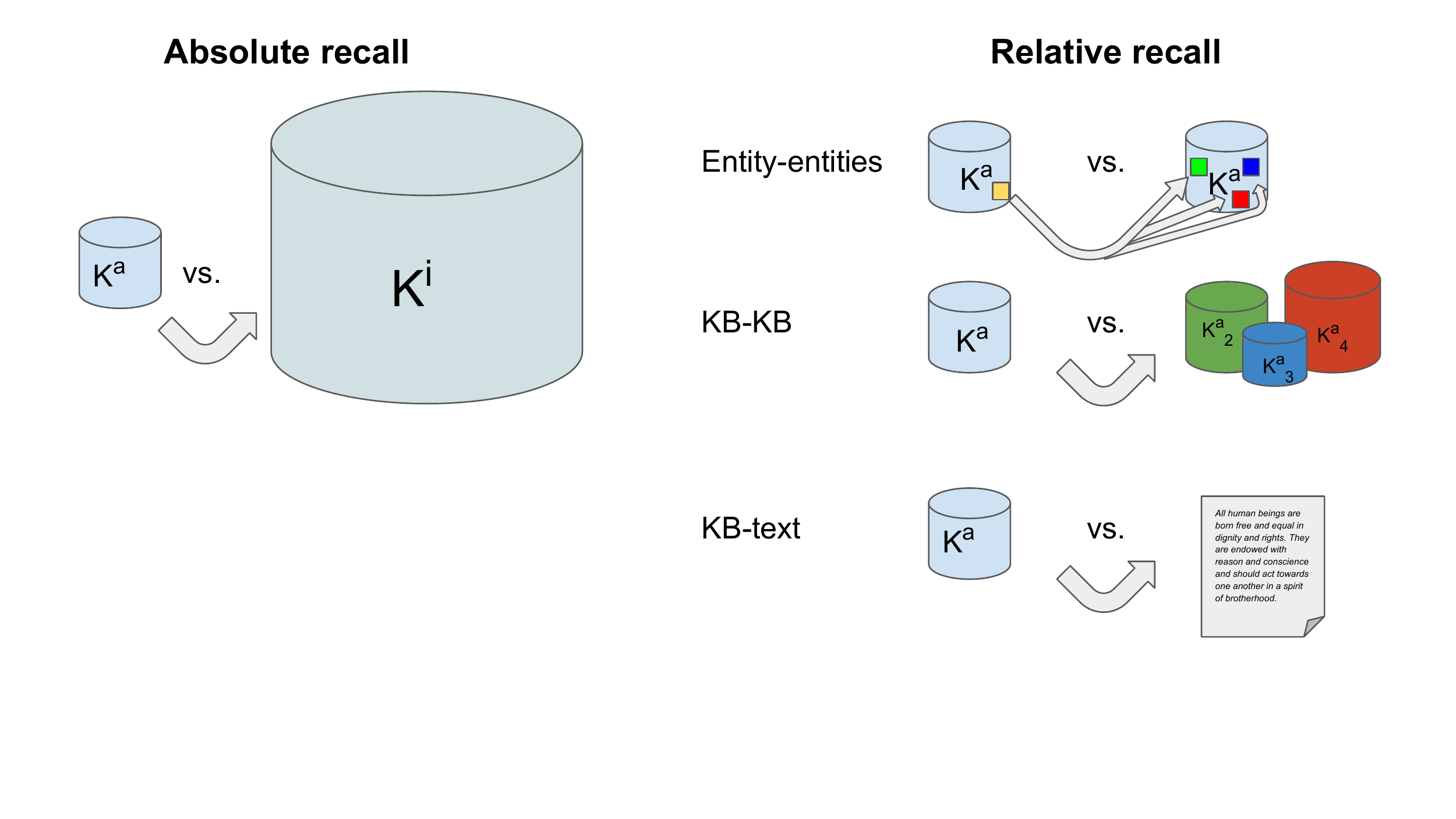}
    \caption{Difference between absolute recall (left) that was discussed in previous sections, and various notions of relative recall (right). 
}
   \label{fig:relrecalloverview}
\end{figure}

So far, the yardstick for recall/completeness has been the real world: \textit{How many of the entities (or predicates or statements) of the domain of interest have been captured?} While this generally is a meaningful target, in some cases, the notion is not well-defined, or not informative. In the following, we look at alternative formulations, by relaxing the absolute yardstick into a relative one:  \textit{Relative to other entities/resources/use cases, how much is the KB's recall?} 
We look at this problem in three variants: (i) recall within the same KB, relative to other entities, (ii) recall relative to other resources like KBs or texts, and (iii) recall relative to the extrinsic use case of question answering (see Figure \ref{fig:relrecalloverview}).

\subsection{Entity Recall Relative to other Entities}
\label{subsec:entity-entity-recall}

We define the problem of relative entity recall as follows: 

\problem{Relative Entity Recall Problem}{An entity $e$, and a KB $K$}{Determine the recall of $K$ for statements about $e$, relative to related entities in $K$.}



\noindent
For example, one may ask how complete Wikidata's information on \textit{Albert Einstein} is, relative to similar entities.
The instantiation of this problem has two major components:
\begin{enumerate}
    \item How are related entities defined? 
    \item How is recall quantified and compared?
\end{enumerate}
Entity relatedness is topic with much history in data mining, and a wide range of text-based, graph-based, and embedding-based similarity measures exists (see Section~\ref{sec:negation} and e.g., \cite{witten2008effective,entity-relatedness}). 
Similarly, recall can be quantified in a variety of ways, for instance, via the number of statements, predicates, inlinks, outlinks, etc. We shall now see different approaches to both problems. 

\paragraph{Relative recall indicators.} One of the first approaches to the problem of relative entity recall is Recoin (\underline{Re}lative \underline{co}mpleteness \underline{in}dicator)~\cite{ahmeti2017assessing,balaraman2018recoin}. It extends the entity page of Wikidata with a traffic-light-style recall indicator, indicating how comprehensive the information is compared to related entities. Its definition of relatedness is class-driven: \textit{Paris} would be compared with other capital cities, \textit{Radium} with other chemical elements, Albert Einstein with other Physicists (treating Wikidata's occupations as pseudo-classes). Its quantification of recall follows a simple frequency aggregation: In each class, the top most frequent properties are computed. For instance, for capitals, the most frequent \textit{predicates} are \textit{country} (99\%), \textit{coordinate location} (97\%), and \textit{population} (82\%). Then, the frequencies of the top-5 absent properties are added up for the given entity, and the sum is compared with 5 global discrete thresholds, to arrive at a final traffic light color. 

\begin{figure}[t]
    \centering
    \includegraphics[width=0.8\columnwidth]{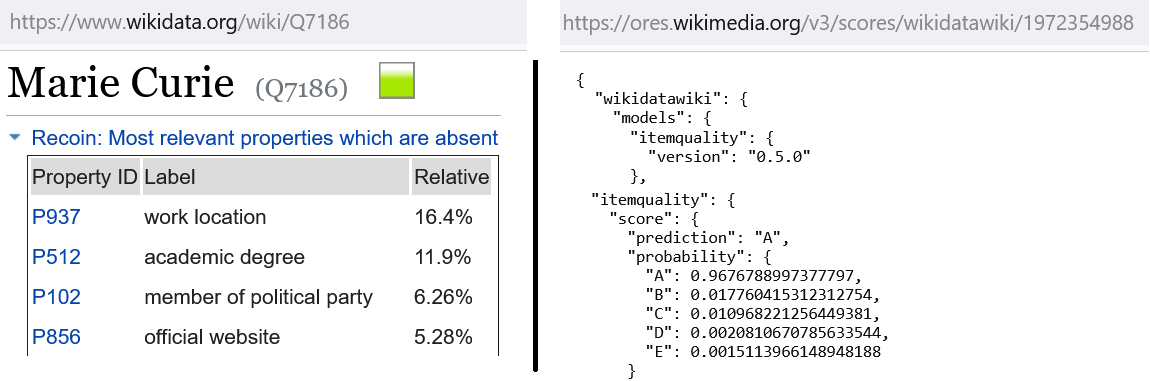}
    \caption{Illustration of Recoin (left) and ORES (right) outputs for \textit{Marie Curie} on Wikidata. Recoin outputs show class frequencies of absent properties, while ORES quality classes A-E are computed using a supervised regression model.}
    \label{fig:recoinvsores}
\end{figure}

A second approach to the problem of relative recall is provided by Wikimedia's ORES machine learning platform~\cite{ores}. The \textit{Wikidata item quality module} specifically assigns probabilities to entities belonging to one of 5 quality levels (A to E), subsuming, besides recall, several other quality dimensions such as completeness, references, sitelinks, media quality. The scores are also relative in the sense that good item quality has no inherent definition, but can only be understood in comparison with other good/bad items. ORES apparently employs supervised machine learning, based on a combination of content embeddings and latent features, yet its concrete working is only partially documented.
We show example outputs of Recoin and ORES for \textit{Marie Curie}, as of September 2023, in Figure \ref{fig:recoinvsores}.

\paragraph{Property ranking.}
Underlying both Recoin and ORES is the question of how to decide whether, for a given entity, a certain predicate is expected to be populated. This is referred to as the problem of \textit{property ranking}. Given an entity like \textit{Marie Curie}, is it more relevant that her doctoral advisor is recorded, or the sports team she played for? 

Property ranking relates to the problem of statement ranking \cite{relevance-facts-ziawasch,bast:relevance:scores:triples}, though the absence of an object value makes the problem harder.
The frequency-based ranking, like the one used in Recoin~\cite{balaraman2018recoin}, provides a reasonable baseline, yet frequencies are only a proxy for relevance. For example, the most frequent properties typically concern generic biographic statements like place of birth, date of birth, while interesting aspects like scientific discoveries, awards, or political affiliations are expressed less frequently. 

Property ranking for relative recall assessment has thus been advanced in several ways: In \cite{razniewski2017doctoral}, text-based predictive models are proposed, which are trained on Wikipedia descriptions of entities, and that describe how likely a textual description is accompanied by a certain structured property in Wikidata. In an evaluation of pairwise property relevance predictions, this approach achieves up to 67\% agreement with human annotators. Gleim et al.\ introduce SchemaTree~\cite{gleim2020schematree}, a trie-based method for capturing property frequencies in the existing data, taking into account not just individual frequencies, but also frequencies of combinations, and fallbacks to base cases for rare combinations. 
Issa et al.\ rely on association rule mining \cite{issa2017assessing}.
Luggen et al.~\cite{luggen2021wiki2prop} propose a method based on multimodal Wikipedia embeddings, taking into account multilingual article descriptions as well as pictures. They train a multi-label neural classifier on these embeddings, for the task of predicting currently present properties in Wikidata, and find that this significantly outperforms previous approaches. 

\paragraph{Analysis.}
The discussed approaches mostly treat entities as sets of properties - distinguishing just whether a property is present or not, but not taking into account how many values are present. For example, even though one award for \textit{Einstein} is listed in a KB, the KB could be missing many others. Disregarding this aspect is pragmatically motivated by the partial completeness assumption (see Section \ref{sec:foundations}): the presence of one value per property likely implies presence of all values. At the same time, extensions towards explicit regard for multi-valued properties would be desirable. A second limitation is inherent to all relative measures: Relative measures may appear good even for bad items, if the comparison is even worse, and vice versa. There is no firm link to the notion of recall from the previous sections, as, even if an entity is well-covered compared with related entities, no strong deductions about its recall, relative to the real world, are possible.

\subsection{KB Recall Relative to other Resources}
\label{subsec:relative:kb-kb}

There are several other categories of resources that can be used for KB recall assessment. A relatively straightforward comparison are other KBs, while texts and elicited human associations provide different foci, on recall w.r.t. information typically conveyed in texts, and w.r.t.\ information that humans spontaneously associate with concepts.

\paragraph{KBs to KBs.} One way to compare KB recall is by comparing their size. Virtually every KB project compares itself with other resources in terms of size, typically counting the number of entities, classes, and statements (for one example, see Table 1 in the Knowledge Vault paper \cite{knowledgevault}). A more fine-grained analysis was performed in \cite{ringler2017one}. For 25 important classes, considerable variance was found, even on KBs derived from the same source, like DBpedia and YAGO. Variances were in part explained by different modelling approaches (e.g., Wikidata's low number of politicians explained by modelling a class via properties). However, size comparisons do not capture whether or to which degree content from one KB is contained in the other -- it may well be that resources have different foci, and that the larger ones still have limited recall w.r.t.\ smaller ones.

To avoid such issues, we can look at the fraction of entities from one contained in the other. The work in \cite{ringler2017one} analyzed that as well, matching entities via simple String distance functions. Merging KBs generally provided potential for increasing recall even for the already bigger resources, because of differences in focus.

\paragraph{KB vs. text.}
Texts are a prime mode of knowledge storage and sharing, and it is natural to ask how well KBs recall information from texts. Notably, KBs were born in part out of the shortcomings of texts in terms of structuring information, so it is very interesting to investigate to which degree the higher structure of KBs comes at a loss of recall.
Moreover, a principled methodology to compare KB recall w.r.t.\ texts enables a comparison on a much wider set of domains and subdomains than the structured comparison in the KB-to-KB setting,
as texts are available in much bigger abundance than structured resources. Various approaches to estimating KB recall w.r.t.\ texts exist.

In the Aristo TupleKB project~\cite{mishra2017domain}, recall of an automatically constructed science KB is compared with information contained in scientific texts. For this recall assessment, Mishra et al.\ assemble a corpus of 1.2M sentences from elementary science textbooks, Wikipedia, and dictionaries, from which they automatically extract relational statements. They can then quantify the ``science recall'' of a KB by measuring which fraction of these statements is contained in a KB. This is done for 5 KBs (WebChild, NELL, ConceptNet, ReVerb-15M, TupleKB). Since predicate names in the KBs vary, this analysis is restricted to 20 general predicates, which are manually mapped to a wider set of surface names. Also, for subject and object match, only headwords are considered. Under these relaxed conditions, the science recall of these KBs is found to be between 0.1\% and 23.2\%.

A related analysis is provided in the context of the OPIEC project \cite{gashteovski2018opiec}, a corpus of open statements extracted from running an open information extraction system on the entire English Wikipedia. In \cite{gashteovski2020aligning}, the authors analyze the relation of OPIEC with DBpedia, in particular, to which degree OPIEC statements are expressible in DBpedia, and to which degree they are actually expressed. Since open information extraction provides a plausible approximation of relational statements contained in text, this evaluation can be seen as a relative recall evaluation of DBpedia w.r.t. Wikipedia. They provide several important insights: (i) 29\% of open statements can be fully expressed in DBpedia, 29\% partially (e.g., with a more specific or more generic predicate), 42\% of open statements are not KB-expressible.\footnote{This aspect can also be considered another dimension of recall, called \textit{schema recall}. This aspect is as also studied in \cite{nakashole2012patty}, where the authors find that popular KB schemata contain between 13-126 predicates of a sample of 163 predicates found in a few Wikipedia articles.} (ii) Adding more complex constructs, like conjunctions or even existential quantification into the KB increases its recall potential. (iii) When it comes to measuring the actual recall, they find that 18\% of open statements are fully present in DBpedia, 23\% partially present, 59\% not at all.

In the temporal dimension, KB vs.\ text recall has been studied using the distant supervision assumption: Assuming that hyperlinked entities on Wikipedia pages are relevant to the page subject, how does their recall in Wikidata change over time? Wikidata's revision history allows to study this longitudinal, allowing to observe a generally steady increase in recall since from 2012 to 2020 \cite{razniewski2020structured}.
 
 
\paragraph{KB vs. human associations.}
Besides other KBs and texts, an interesting resource for relative recall assessments are humans directly. How well do KBs recall statements that humans spontaneously associate with concepts? This framing of recall has been prominent in recent commonsense knowledge base construction (KBC) projects \cite{romero2019commonsense,nguyen2021materialized,ascentpp}.
While the former project directly queried human crowdworkers (``What comes to your mind when you think of lions?''), the latter two projects relied on the CSLB property norm corpus \cite{devereux2014centre}, a large dataset of concept associations collected in the context of a psychology project. Evaluations following this scheme typically use embedding-based heuristic matching techniques to judge whether a test KB contains a reference statement subject to possible minor wording differences. Recall is typically found to be in the order of 5\%-13\%, showing that there is still a substantial gap in how well current commonsense KB
projects cover human knowledge.

\paragraph{Analysis.}
We have discussed three ways of estimating KB recall relative to other resources. Common among them is the challenge of how to compare pieces of knowledge: Statements in the reference resource may be differently worded or ambiguous, making finding a statement of equivalent semantics nontrivial. Moreover, semantic equivalence is necessarily an imprecise concept, requiring somewhat arbitrary decisions about the maximally accepted dissimilarity, as well as difficult technical decisions on how to actually measure semantic relatedness. A common technical solution currently is to represent the two statements to be compared in latent embedding space (e.g., \cite{nguyen2021materialized} uses S-BERT \cite{reimers2019sentence}), then use similarity metrics like cosine distance to decide whether two candidates are considered equivalent. Note that this problem even occurs in the most structured setting, KB-vs.-KB recall, since even if subjects and objects are disambiguated, predicate names are typically textual strings, like \textit{worksAt, employer, affiliatedWith}. 

Each of the mentioned comparisons serves a different purpose: Evaluating KB recall relative to other KBs should definitely be attempted whenever similar-domain KBs are available, since data integration from structured sources is a comparatively simple task, with considerable potential for improving recall.
Evaluating KB recall relative to text can serve two purposes: Towards improving KB recall, for text-extracted KBs, recall evaluation relative to text can serve as a guidance on how much information is lost in the text extraction process. This can serve as a guidance on where to improve the process. On the other hand, such an evaluation can also help to understand a KB's potential and limitation for downstream use cases (e.g., Wikidata is generally suited for computing statistics about age/gender/profession statistics about US senators, but bad for judging the quality of their governance). This is especially relevant when KBs are one of several options to power a downstream use case, as is, for example, often the case for question answering (see also Section~\ref{sec:relative-qa}).
KB recall relative to human associations is the most intrinsic dimension.


\subsection{KB Recall Relative to Question Answering Needs}
\label{sec:relative-qa}

Another relative way to evaluate KB recall is by considering a use case, and quantifying to which degree KBs satisfy its data needs. Natural-language \textit{question answering} is arguably a supreme task in knowledge management and NLP, and as such especially suited to illustrate this point.

\paragraph{Static analyses.}
Several papers investigate to which extent KBs allow answering questions from common QA datasets. Most prominently, this happens in scenarios where text-based QA systems and KB-based QA systems are compared. Two especially illustrative comparisons can be found in \cite{oguz2020unik} and \cite{pramanik2021uniqorn}.
The first work analyzes the performance of a state-of-the-art QA system on 4 popular benchmarks of general-world questions (NaturalQuestions, WebQuestions, TriviaQA, CuratedTREC). It finds that KB-based systems can correctly answer between 26\% and 43\% of these queries. It also provides a comparison of KB recall to text recall (see Section~\ref{subsec:relative:kb-kb}), by comparing the previous numbers with the performance of text-based QA systems (finding 45-62\% recall). In other words, texts generally provide higher recall than KBs, although this needs to be weighed against other advantages of structured resources. The work of \cite{pramanik2021uniqorn} analyzes complex queries in  6 QA benchmark datasets (LCQuad 1.0 and 2.0, WikiAnswers, Google Trends, QuALD, ComQA), finding that a state-of-the-art system can answer 10-19\% of these queries by using only KBs, and 18-36\% by only using text. Further variants of this kind of analysis exist, for example for conversational question answering \cite{christmann2022conversational}.

The common insight from these studies is that KBs provide encouraging recall for utilizing them in QA systems, but are far from saturating the query sets, thus often motivating hybrid QA systems that combine KBs and text. At the same time, 
the reported scores typically conflate the recall of the KB and the ability of the system to 
pull out the correct answers from the KB. Since query answering is a heuristic, imperfect process, the intrinsic KB recall for these QA datasets is likely higher.

\paragraph{Predicting recall requirements for the QA use-case.}
An interesting twist to recall analysis is provided by Hopkinson et al.\ \cite{hopkinson2018demand}: Instead of assuming a fixed KB and measuring recall, they define a desired query recall (95\% of queries should be answered by the KB), and ask which content needs to go into the KB to achieve this. This is not an obvious question, because information needs may vary highly depending on the type of entity, and questions do not uniformly target entities and properties. The work originates in an industrial lab (Amazon Alexa), where such a business requirement is plausible. The QA service provider here has the potential to design automated extraction efforts accordingly, or to task paid KB curators to complete specific areas of the KB. Moreover, commercial service providers have access to user query logs, which underlie this technique. On a technical basis, the work represents entities via their class membership, extracts usage frequencies of properties per entity from the query log, and predicts predicate usage patterns on new entities using either a regression or a neural network model. Results indicate that this method can predict required properties with good accuracy. Furthermore, demand-weighted requirements can be lifted to the level of the whole KB, based on usage data about which entities are queried how often. For a non-representative set of entities in the Alexa KB, the authors find that 58\% of the predicates needed to arrive at the 95\% query-answerability goal are currently in the KB.

\paragraph{Longitudinal development.}
\cite{razniewski2020structured} analyzes how KB utility has changed over time. The authors select questions from three search engine logs (AOL, Google, Bing queries), then use human annotation to find out the earliest time at which a KB (Wikidata and DBpedia) could answer these questions. For example, the question \textit{``Where is Italian Job filmed?''} can be answered by Wikidata since October 15, 2015, when the property \textit{filming location} on the entity \textit{Italian Job} was added. Plotting the number of answerable queries over time turns out to show a steady increase for the time period from 2003 to 2020, with only minor slowing in recent years.

\subsection{Summary}

\begin{table}[]
\begin{adjustbox}{width=\textwidth,center}
\begin{tabular}{p{3.2cm}p{3cm}p{1.8cm}p{5cm}p{4.5cm}}
\toprule
\textbf{Focus of relative recall} & \textbf{Aspect} & \textbf{Works} & \textbf{Strength} & \textbf{Limitations} \\ \midrule
\multirow{2}{*}{Entity vs. other entities} & Relative entity recall & \cite{balaraman2018recoin,ores} & Enables ranking/priorization without knowledge about reality & Struggles if many properties are optional/if only parts of values are present\\
 & Determining most relevant predicates per entity & \cite{razniewski2017doctoral,gleim2020schematree,luggen2021wiki2prop} & Produces interpretable suggestions on where to complete KB & Struggles with optional predicates \\ \hline
\multirow{3}{*}{KB to other resources} & KB to KB & \cite{knowledgevault,ringler2017one} & Can give quick suggestions on when to integrate data & Similar-topic KBs often not available \\
 & KB to text & \cite{mishra2017domain,gashteovski2018opiec,gashteovski2020aligning,razniewski2020structured} & Can help identifying issues in text extraction, or help in choices between KB or text-based downstream applications & Threshold of when textual statement is covered in KB not obvious \\
 & KB to human associations & \cite{romero2019commonsense,nguyen2021materialized,ascentpp} & Highest aspiration of all evaluations & Practical implications not clear \\
 \hline
\multirow{3}{*}{KB to QA use cases} & Counting \#queries answered by KB-QA system &  \cite{oguz2020unik,pramanik2021uniqorn,christmann2022conversational} & Gives tangible insights into how well KB feeds a use case & Difficult to disentangle KB recall and QA system performance \\
 & Predictive QA recall & \cite{hopkinson2018demand} & Allows to predict content needed in KB for meeting a use case requirement & Requires substantial query logs \\
 & Longitudinal development & \cite{razniewski2020structured} & Shows that KBs have steadily improved for QA & Relies on heuristic matches \\ \bottomrule
\end{tabular}
\end{adjustbox}
\caption{Summary of relative recall assessment. 
}
\label{tab:relativerecallsummary}
\end{table}

Recall estimation in absolute terms is generally a difficult task. This section has provided a pragmatic alternative, showing how to measure KB entity and statement recall relative to other KBs, other resources, and QA use cases.

We summarize the insights from this section in Table~\ref{tab:relativerecallsummary}. Each of these approaches comes with advantages and disadvantages, mostly stemming from challenges in how to measure relative recall, and the potential of systematic omissions in the reference.

\textit{Relative entity recall} is arguably the easiest to analyze, and the quantification is comparatively easy, as matching predicates within a KB is simple. Thought is still required in the definition of the reference entities, e.g., comparing Einstein with other Physicists may give very different results, than comparing him with other violin players. Furthermore, relative entity recall is prone to systematic gaps in the KB, e.g., if the predicate \textit{known for} is entirely absent from the KB, then there is no way to suggest it for Einstein either.

\textit{Relative recall to other resources} provides a more external view of a KB, and especially when choosing text as reference, provides potential for many interesting analyses (e.g., recall of Einstein's KB entry can be computed w.r.t.\ Wikipedia, w.r.t.\ Simple Wikipedia, w.r.t.\ a biographical book, etc.). At the same time, quantifying recall w.r.t.\ external resources is more challenging, as it requires dealing with schema matching (KB-KB setting), or imprecise and ambiguous predicate and entity surface forms (KB-text setting).

\textit{Relative recall w.r.t.\ use cases} provides clear metrics of how a KB is faring downstream, and this is advantageous if a KB is built with a primary use case in mind. This strength is however often also a challenge, since often, KB construction is a longitudinal, cross-functional endevaour \cite{machine-knowledge-survey}, where use cases are moving targets.


\section{Discussion and Conclusion}

We conclude this survey with \begin{revision}
a discussion of the temporal dimension of recall (Section~\ref{subsec:temporal}), the impact of large language models (Section~\ref{subsec:llms}),\end{revision}\ recommendations on how to perform KB recall assessment (Section~\ref{subsec:recipe}), a set of take-home lessons (Section~\ref{subsec:takehome}), and list of challenges for future research (Section~\ref{subsec:openchallenges}).

\begin{revision}

\subsection{The Temporal Dimension of Completeness and Recall}
\label{subsec:temporal}

Reality continuously changes: people who were once presidents lose that role, people marry or divorce, and people who once did not have a death date may obtain one. 
Knowledge bases follow this course, and are usually updated.  
In this section, we discuss the impact that these changes have on the tasks of completeness and recall estimation.

Many KBs grow steadily. Wikidata, for instance, contained 20M statements in 2015, but 1.2B statements in 2023, hinting at a substantial increase of its recall. Reality, on the other hand, evolves as well, which may mean that areas formerly complete might become incomplete later on. In this section, we discuss formalisms for temporal annotations, methods for extraction and extrapolation, and observational studies concerning recall trends.

\paragraph{Formalisms for Temporal Annotations.}
Completeness and cardinality statements can be extended with information on their temporal validity. Darari et al. \cite{darari2018completeness}, for instance, introduced the notion of \emph{time-stamped completeness statements}. These add a ``latest validity'' date to a completeness statement (which is typically, but not necessarily, the date of their creation). Examples are statements like \textit{``Nobel Prize winners are complete until 2023''}, or \textit{``XYZ's publications are complete until 2018''}.
Similarly, Arnaout et al.~\cite{ARNAOUT2021100661} extended negative statements inference with a notion of temporal prefix, allowing to conclude, for instance, that unlike her 7 predecessors, the German chancellor Angela Merkel was not male.

In many KBs, temporal annotations are also used for positive statements. A typical statement, in Wikidata, for instance, qualifies that Albert Einstein's German citizenship ended in 1933, or that he received the Nobel Prize in Physics in 1921.\footnote{\url{https://www.wikidata.org/wiki/Q937}} Annotating completeness and recall statements with temporal qualifications appears therefore quite natural.

\paragraph{Extracting and extrapolating time information.}
For completeness and recall information that is text-extracted, a reasonable baseline is to consider the extraction time as the latest validity time. Finer-grained annotations are possible, for instance, by considering document creation time metadata, or temporal expressions in the actual text, for estimating the latest (certain) validity time \cite{heideltime}.

Given time-annotated completeness or recall metadata, an orthogonal question is how to interpret it after its latest validity time. For instance, if publications by Albert Einstein were complete in 2018, it is reasonable to assume that the same is the case in 2023. Yet no such conclusion should be drawn for books \textit{about} Albert Einstein.
Drawing this distinction means entering the realm of predicting the temporal stability of knowledge, a problem that comes with a dependence on domain knowledge, and which appears under-explored for both structured knowledge \cite{decay,dikeoulias2019epitaph} and text \cite{jatowt1}.

\paragraph{Recall trends.}
KBs like Wikidata, YAGO, and DBpedia are under active development, and one may wonder how their recall evolves. In Section \ref{sec:relative-qa}, we discussed work that analyzed recall relative to QA needs, and Wikipedia content, for DBpedia and Wikidata \cite{razniewski2020structured}. Remarkably, relative to fixed information needs (QA logs from the 2000s, a snapshot of Wikipedia pages), recall had constantly increased from 2003-2020, with only a minor slowdown in recent years. However, reality and information needs also evolve over time, and hence it remains an open question of whether KBs are faster in capturing reality, or reality develops faster than KBs can represent it.

\subsection{The Impact of Large Language Models}
\label{subsec:llms}

Recently, large language models (LLMs) such as Bert \cite{bert}, (chat-)GPT~\cite{radford2019language}, and LLaMA~\cite{touvron2023llama} have significantly impacted natural language processing. This impact has extended to knowledge-intensive tasks, and specifically also to knowledge bases \cite{pan2023large,kbs-and-llms}. Although completeness and recall research has yet to capitalize on these advances, there are several ways by which LLMs are likely to impact this area.

\paragraph{Indirect impact: KBs with higher recall.} LLMs support many steps in the KB construction pipeline, thereby enabling the construction of bigger KBs, that consequently have higher recall \cite{alivanistos2022prompting,veseli2023evaluating}. In particular, LLMs can be used both for direct knowledge prediction, or in conjunction with retrieved text, where they improve over existing textual relation extraction methods \cite{mallen2023not}.

\paragraph{Direct impact: Easier linking of text and structured modalities.} Linking textual statements with structured statements is a problem that affects several of the discussed methodologies, most notably peer-based inferences (see Section \ref{sec:negation}, where existing statements need to be matched with statements on peers, and with textual evidence), and the KB-to-text relative recall assessment (Section \ref{subsec:relative:kb-kb}). Due to their significant capabilities in latently representing and matching assertions in different formulations and representations, we may expect advances on these parts soon. 

\paragraph{Direct impact: Conversational maximes.} LLMs are especially strong in generalizable linguistic tasks that do not require instance-specific knowledge. In the context of recall assessment, one such task concerns the estimation of whether a textual extraction context implies completeness, or not (see Section~\ref{sec:recall-facts}). For example, chatGPT can give convincing answers to the following examples that come with very subtle linguistic nuances: 
\begin{center}
\fbox{%
\parbox{0.9\textwidth}{%
\textbf{User:} Text entailment: Please estimate how likely the first sentence entails the second.\\   
\mbox{\hspace{0.85cm}} Premise: John brought his daughter Mary to school.\\    
\mbox{\hspace{0.85cm} }Hypothesis: John has only one child.\\

\textbf{chatGPT}: [..] high likelihood.\\

\textbf{User:} 
Text entailment: Please estimate how likely the first sentence entails the second.\\   
\mbox{\hspace{0.85cm} }Premise: John also brought his daughter Mary to school.\\
\mbox{\hspace{0.85cm} }Hypothesis: John has only one child.\\

\textbf{chatGPT}: [..] does not entail.
}%
}
\end{center}

\paragraph{Direct impact: Generating metadata.} 
Mirroring approaches that directly prompt LMs for statements \cite{alivanistos2022prompting,veseli2023evaluating}, LLMs can in principle also be prompted for completeness or recall metadata. An example is shown next.
\begin{center}
\fbox{%
\parbox{0.9\textwidth}{%
\textbf{User:} According to Wikidata, Barack Obama has two children, Malia and Sasha. Are these truly all?\\

\textbf{chatGPT}: Yes, according to publicly available information, Barack Obama has two daughters [..].\\

\textbf{User:} According to Wikidata, Albert Einstein has published 75 scholarly articles. Are these truly all?\\

\textbf{chatGPT}: No, the number of scholarly articles published by Albert Einstein is estimated to be around 300. [..]
}%
}
\end{center}

\ \\
In the example above, chatGPT gives correct answers to both questions, confirming completeness for Barack Obama's children, and rejecting completeness for Einstein's publications, while also giving recall information (75/300).
However, metadata obtained in this way itself comes with huge credibility issues. LLMs are known to confabulate, especially for long-tail topics, so without a proper sourcing of answers, such derivations are likely difficult to utilize downstream. Sourcing LLM generations in texts is difficult for principled reasons, and an open research challenge \cite{zhao2023explainability}.
One exception to this general issue may be negative commonsense knowledge (cf.\ Section \ref{subsec:negation:csk}). Here, correctness depends anyway much on context, and requirements in correctness are balanced by requirements for saliency. In this case, LLMs can reasonably generate interesting negation candidates \cite{hiba-llms-negation}. 

\subsection{Recommendations  for KB Recall Assessment}
\label{subsec:recipe}

\end{revision}

We have aimed for a balanced coverage of approaches so far, which may leave practically-minded readers wondering how they could best approach a specific KB recall assessment problem. In the following, we give more concrete suggestions on how we think a sensible order of approaches to specific problems could be. We distinguish three settings: (i) open-ended ab-initio KB construction, where recall-awareness can be intertwined with construction efforts, (ii) use-case driven construction, where efforts can be directly matched with use case metrics, and (iii) KB curation, where an existing KB shall be evaluated. 

\paragraph{Setting 1: Open-ended ab-initio KB construction.}
For open-ended ab-initio KB construction, i.e., the novel construction of a KB intended for broad use, our suggestion is to intertwine the data acquisition process with metadata acquisition. Concretely, if the KB content is text-extracted, we suggest to use the text recall estimation techniques from Section~\ref{sec:recall-facts}, to annotate extractions with confidences in completeness. If the data is created by human authors, we suggest to augment the data authoring tools with fields for metadata collection, e.g., checkboxes that allow authors to note when they finished recording a topic, similar as in Cool-WD \cite{prasojo2016managing}.

\paragraph{Setting 2: Use-case-driven ab-initio KB construction.}
For KB construction driven by a specific use case, our suggestion is to organize the recall assessment via metrics derived from the use, as discussed in Section~\ref{sec:relative-qa}. Initially, one should derive a profile of queries to be answered by the KB, e.g., by sampling from the use case. Where the sample's breadth is limited, interpolation should be used to derive a broader profile \cite{hopkinson2018demand}. Efforts towards KB population can then be evaluated against this query profile, i.e., for each specific population technique or domain, one could compare cost and benefit, and prioritize accordingly.

\paragraph{Setting 3: KB curation.}
In settings where existing KBs shall be evaluated, we suggest to first check for the existence of KB-internal cardinality information, as discussed in Section \ref{subsec:cardinalityinkb}. Next, if high-quality texts like Wikipedia are available, we suggest to exploit textual cardinality assertions, preferably with simple template-based extraction, as in Section \ref{subsec:cardinalityintext}. Relative recall, in particular by comparing entities inside the KB (see Sec.~\ref{subsec:entity-entity-recall}), will also help in spotting gaps. Statistical properties such as the ones we have discussed in Section \ref{sec:predictive} should only be used once all other options have been exhausted, because they are least reliable. 

\subsection{Take-home Lessons}
\label{subsec:takehome}

The key takeaways from this survey are:
\begin{enumerate}
    \item \textbf{KBs are incomplete}: Despite the long history of the fields of KB construction, Semantic Web, and Information Extraction, the construction of a general-world knowledge base is an inherently fuzzy and evolving task. Therefore, such KBs will always be incomplete, and one has to be able to deal with this incompleteness, rather than hoping that it will disappear (Section~\ref{sec:foundations}).
    \item \textbf{KBs do hardly contain negative information (but should)}: Negative information is very useful for downstream tasks, 
    but regrettably underrepresented in current KBs. Selective materialization of interesting negations can significantly enhance the utility of KBs (Sections \ref{sec:foundations} and \ref{sec:negation}). 
    \item \textbf{Predictive techniques work for a surprising set of paradigms}: Besides classical supervised prediction, there are statistical properties like number distributions, sample overlap, and density invariants that enable recall prediction even without typical training (Section \ref{sec:predictive}).
    \item \textbf{Count information is a prime way to gain insights into recall}: Count information provides the most direct way to recall assessment, and it can be found both in existing KBs, and in text (Section \ref{sec:cardinalities}).
    \item \textbf{Salient negations can be heuristically materialized}: Although negative knowledge is quasi-infinite, heuristics for materializing relevant parts can significantly complement positive-only KBs (Section \ref{sec:negation}).
    \item \textbf{Relative recall is a tangible alternative to absolute notions}: Comparing KB entities with other KB entities, external resources, or use case requirements provides a valuable second view on KB recall (Section \ref{sec:relativerecall}).
\end{enumerate}

\subsection{Challenges and Opportunities}
\label{subsec:openchallenges}

In this final section, we sketch some of the open challenges that remain to be addressed
towards fully understanding KB recall, pointing out opportunities for original and potentially impactful research.
\begin{enumerate}
    \item \textbf{Developing high-accuracy recall estimators.} Most of the estimators presented in this survey are proofs-of-concept, tested only in limited domains, or under very specific assumptions. Building practically usable high-accuracy estimators, possibly by combining several complementary estimation techniques, remains a major open challenge. 
    \item \begin{revision}
        \textbf{Exploiting recall estimates for value-driven KB completion.} Despite their obvious connection, research on recall estimation and KB completion has so far evolved largely independently. Quantifying the value of knowledge (as in Section \ref{sec:relative-qa}), and defining prioritization strategies for recall improvement that maximize the value of the available knowledge \cite{hopkinson2018demand,decay}, are great opportunities for practically impactful research.\end{revision}
    \item \textbf{Estimating the recall of pre-trained language models.} Knowledge extraction from pre-trained language models has recently received much attention \cite{lama}, yet it remains unclear to which degree this approach can yield knowledge for multi-valued, optional, and long-tail predicates \cite{jiang2021can,lm-kbc-challenge}. Systematically measuring the recall of language models, and comparing it with structured KBs, is an open challenge.
    \end{enumerate}
%
Knowledge bases have received substantial attention in recent years, and while precision is usually in the focus of construction, understanding their recall remains a major challenge. In this survey we have systematized major avenues towards KB recall assessment, and outlined practical approaches and open challenges. 
We hope this survey will inspire readers to reflect on KB quality from a new angle, and lead to more KB projects that systematically record and reflect on their recall.

\bibliographystyle{ACM-Reference-Format}
\bibliography{updatedrefs}

\end{document}